\documentclass[10pt,journal,compsoc]{IEEEtran}

\ifCLASSOPTIONcompsoc
  \usepackage[nocompress]{cite}
\else
  \usepackage{cite}
\fi
\ifCLASSINFOpdf
  \usepackage[pdftex]{graphicx}
  \graphicspath{{figures/}}
  \DeclareGraphicsExtensions{.pdf,.jpeg,.png,.jpg}
\else
\fi
\usepackage{tikz}
\usetikzlibrary{positioning}
\usetikzlibrary{shapes}
\usetikzlibrary{plotmarks}

\usepackage{amsmath}
\DeclareMathOperator*{\argmin}{arg\,min}

\usepackage{algorithmic}
\ifCLASSOPTIONcompsoc
  \usepackage[caption=false,font=footnotesize,labelfont=sf,textfont=sf]{subfig}
\else
  \usepackage[caption=false,font=footnotesize]{subfig}
\fi
\usepackage{hyperref}

\usepackage{multirow}

\newtheorem{definition}{Definition}

\begin{document}
%
\title{A General Framework for the Recognition\\ of Online Handwritten 
Graphics}
%
%
%
%

\author{Frank Julca-Aguilar,
        Harold Mouch\`{e}re,
        Christian Viard-Gaudin,
        and~Nina S. T. Hirata
\IEEEcompsocitemizethanks{\IEEEcompsocthanksitem Frank Julca-Aguilar and Nina S. T. Hirata
are with the Department of Computer Science, Institute of Mathematics and 
Statistics, University of S\~{a}o Paulo, Brazil.\protect\\
E-mail: \{faguilar, nina\}@ime.usp.br
\IEEEcompsocthanksitem Harold Mouch\`{e}re and Christian Viard-Gaudin 
are with Institut de Recherche en Communications et 
Cybernétique of Nantes, University of Nantes.\protect\\
E-mail: \{christian.viard-gaudin, harold.mouchere\}@univ-nantes.fr}
}

%
%

\markboth{Submitted on September~2017}%
{Julca-Aguilar \MakeLowercase{\textit{et al.}}: A General Framework
  for the Recognition of Online Handwritten Graphics}
%



\IEEEtitleabstractindextext{%
\begin{abstract}

We propose a new framework for the recognition of online handwritten
graphics. 
Three main features of the framework are its ability to
treat symbol and structural level information in an integrated way,
its flexibility with respect to different families of graphics, and
means to control the tradeoff between recognition effectiveness and
computational cost. We model a graphic as a labeled graph generated
from a graph grammar. Non-terminal vertices represent subcomponents,
terminal vertices represent symbols, and edges represent relations
between subcomponents or symbols.
We then model the recognition
problem as a graph parsing problem: given an input stroke set, we
search for a parse tree that represents the best interpretation of the
input. Our graph parsing algorithm generates multiple interpretations
(consistent with the grammar) and then we extract an optimal
interpretation according to a cost function that takes into
consideration the likelihood scores of symbols and structures. The
parsing algorithm consists in recursively partitioning the stroke set
according to structures defined in the grammar and it does not impose
constraints present in some previous works (e.g. stroke ordering).
By avoiding such constraints and thanks to the powerful
representativeness of graphs, our approach can be adapted to the
recognition of different graphic notations.
We show applications to the recognition of mathematical expressions
and flowcharts. Experimentation shows that our method obtains
state-of-the-art accuracy in both applications.
\end{abstract}

\begin{IEEEkeywords}
Graphics recognition, online handwriting recognition, graph parsing,
mathematical expression, flowchart.
\end{IEEEkeywords}}

\maketitle

\IEEEdisplaynontitleabstractindextext

%
\IEEEpeerreviewmaketitle

\IEEEraisesectionheading{\section{Introduction}\label{sec:introduction}}
%
%
%
%

\begin{figure}[htb]
\begin{center}
\resizebox{0.65\linewidth}{!}{\begin{tikzpicture}[x=1, y=-1,scale=1.2]


\draw[gray, ultra thick] plot coordinates {(238, 135) (236, 134)(236, 135)(236, 134)(235, 134) (243, 136) (247, 136) (252, 136) (258, 136) (264, 136) (281, 135) (286, 135) (289, 135)};

\draw[gray, ultra thick] plot coordinates {(237, 136) (243, 139) (247, 141) (251, 142) (257, 143) (261, 145) (266, 146) (271, 147) (274, 149) (277, 152) (276, 153) (274, 155) (270, 157) (266, 159) (260, 162) (254, 165) (243, 171) (238, 174) (235, 177) (233, 180) (240, 180) (245, 180) (251, 180) (258, 180) (265, 180) (273, 181) (280, 181) (286, 182) (291, 182) (295, 182) (299, 182) (300, 182)};

\draw[gray, ultra thick] plot coordinates {(255, 86) (255, 87) (256, 94) (256, 98) (257, 103) (259, 108) (260, 112) (260, 115) (260, 118) (261, 119) (260, 118) (260, 117) (261, 109) (262, 105) (263, 103) (265, 101) (266, 100) (267, 100) (267, 99) (268, 100) (267, 103) (265, 107) (269, 113) (271, 115) (273, 116) (276, 118) (277, 118)};

\draw[gray, ultra thick] plot coordinates {(229, 207) (225, 203) (224, 201) (223, 199) (223, 198) (223, 199) (223, 200) (224, 201) (225, 213) (225, 214) (226, 214) (226, 213) (227, 204) (227, 200) (228, 197) (229, 195) (231, 194) (233, 194) (236, 197) (237, 200) (238, 203) (238, 206) (238, 209) (239, 211) (239, 212)};

\draw[gray, ultra thick] plot coordinates {(253, 200) (254, 200) (260, 200) (263, 200) (265, 200) (266, 201)};

\draw[gray, ultra thick] plot coordinates {(255, 209) (260, 210) (263, 210) (271, 211)};

\draw[gray, ultra thick] plot coordinates {(277, 206) (280, 205) (282, 203) (284, 201) (287, 199) (289, 197) (290, 194) (292, 193) (292, 192) (293, 192) (293, 205) (293, 208) (292, 211) (292, 213) (290, 215) (289, 216) (287, 216) (286, 216) (287, 216) (294, 215) (297, 215) (301, 215) (304, 215) (306, 215)};

\draw[gray, ultra thick] plot coordinates {(333, 160) (337, 160) (339, 161) (342, 162) (344, 164) (345, 166) (345, 169) (343, 171) (341, 173) (339, 175) (337, 177) (335, 177) (336, 178)};

\draw[gray, ultra thick] plot coordinates {(352, 160) (352, 161) (351, 162) (348, 168) (347, 172) (348, 174) (354, 177) (358, 178) (362, 178) (365, 178) (367, 178)};

\draw[gray, ultra thick] plot coordinates {(373, 183) (374, 185) (374, 187) (373, 188) (374, 189) (374, 188) (374, 186) (376, 180) (377, 178) (379, 177) (380, 177) (382, 178) (383, 185) (383, 187) (384, 189) (384, 191)};

\draw[gray, ultra thick] plot coordinates {(403, 163) (411, 164) (415, 164) (418, 165) (421, 165) (423, 166) (424, 167) (424, 166) (424, 167) (421, 171) (418, 173) (411, 179) (410, 180) (411, 181) (412, 182) (415, 182) (418, 182) (421, 182) (424, 182) (427, 181) (429, 181) (430, 181)};

\draw[gray, ultra thick] plot coordinates {(405, 170) (406, 170) (407, 170) (409, 170) (423, 170) (429, 171) (431, 171)};

\draw[gray, ultra thick] plot coordinates {(432, 189) (432, 192) (432, 195) (433, 196) (433, 197) (434, 198) (434, 195) (435, 193) (436, 191) (437, 188) (439, 187) (440, 186) (443, 187) (444, 189) (446, 192) (447, 195) (448, 198) (448, 200)};

\draw[blue] (238, 135) node[anchor=south] {$3$};

\draw[blue] (237, 136) node[anchor=east] {$1$};

\draw[blue] (255, 86) node[anchor=south] {$2$};

\draw[blue] (219, 207) node[anchor=east] {$4$};

\draw[blue] (253, 200) node[anchor=south] {$5$};

\draw[blue] (255, 209) node[anchor=north] {$6$};

\draw[blue] (277, 206) node[anchor=south] {$7$};

\draw[blue] (333, 160) node[anchor=south] {$8$};

\draw[blue] (352, 160) node[anchor=west] {$9$};

\draw[blue] (373, 183)  node[anchor=east] {${10}$};

\draw[blue] (403, 163) node[anchor=south] {${11}$};

\draw[blue] (405, 170)  node[anchor=east] {${13}$};

\draw[blue] (432, 189) node[anchor=east] {${12}$};

\end{tikzpicture}}

\bigskip

\resizebox{0.8\linewidth}{!}{
%
%
%
\begin{tikzpicture}[->]
  \def\ggDifference{-3em}
  \def\nodetoNodeDistance{4.5em}
  \def\nodetoNodeDoubleDistance{7em}
  \def\nodetoNodeXDominatedDistance{3em}
  \def\nodetoNodeYDominatedDistance{2.5em}
  \def\baseNodeYDistance{4ex}
  \def\supHeight{6em}
  \tikzstyle{vertex}=[ellipse,fill=black!25,minimum size=17pt,inner sep=0pt, draw=black]
  \tikzstyle{vertex1}=[ellipse,darkgray,fill=white,minimum size=17pt,inner sep=0pt, draw=black]
  \tikzstyle{vertex2}=[ellipse,red,fill=white,minimum size=17pt,inner sep=0pt, draw=red]
  \tikzstyle{vertex3}=[ellipse,darkgray,fill=white,minimum height=4em, minimum width=13em,inner sep=0pt, draw=black]
   \tikzstyle{vertex4}=[ellipse,darkgray,fill=white,minimum height=7em, minimum width=18em,inner sep=0pt, draw=black]
  \tikzstyle{boldVertex}=[ellipse,fill=black!45,minimum size=17pt,inner sep=0pt, draw=black]
   \tikzstyle{grammarGraph}=[rectangle, fill=white,minimum height=3em, minimum width=16em, inner sep=0pt, draw=white]
   \tikzstyle{grammarGraph1n}=[rectangle, fill=white,minimum height=3em, minimum width=4em, inner sep=0pt, draw=white]
   \tikzstyle{grammarGraph1Supn}=[rectangle, fill=white,minimum height=\supHeight, minimum width=21.5em, inner sep=0pt, draw=white]
   \tikzstyle{grammarGraph2Supn}=[rectangle, fill=white,minimum height=\supHeight, minimum width=26em, inner sep=0pt, draw=white]
  

  \node[vertex1] (sum) at (0, 0) {$\sum$};
  \node[vertex1, above = of sum] (k) {$k$};
  \node[vertex3, below = of sum] (neqone) {};
  \node[vertex1, below  left = 1.5 cm and 1.2 cm of sum] (n1) {$n$};
  \node[vertex1, right = of n1] (equal) {$=$};
  \node[vertex1, right = of equal] (one) {$1$};
  \node[vertex4, right = 2.2 cm of sum] (xz) {};
  \node[vertex1, above right  = 0.05 cm and 3cm of sum] (x) {$x$};
  \node[vertex1, below right = 0.8 cm of x] (n2) {$n$};
  \node[vertex1, right = 2cm of x] (z) {$z$};
  \node[vertex1, below right = 0.8 cm of z] (n3) {$n$};

  \draw [->,black] (sum) -- (k) node[left,midway] {above};
  \draw [->,black] (n1) -- (equal) node[above,midway] {right};
  \draw [->,black] (equal) -- (one) node[above,midway] {right};
  \draw [->,black] (x) -- (n2) node[right,midway] {subscript};
  \draw [->,black] (x) -- (z) node[above,midway] {right};
  \draw [->,black] (z) -- (n3) node[right,midway] {subscript};
  \draw [->,black] (sum) -- (neqone) node[left,midway] {below};
  \draw [->,black] (sum) -- (xz) node[above,midway] {right};

  


\end{tikzpicture}
 
}
\caption{\label{fig:me_input} Handwritten mathematical expression example. 
Top: A sequence of strokes where the order (indicated by numbers in
blue) is given by the input time. Symbols $\sum$ and $z$ are composed
of non-consecutive strokes. Bottom: The expression is composed of
symbols and several types of spatial relations between them.}
\end{center}
\end{figure}
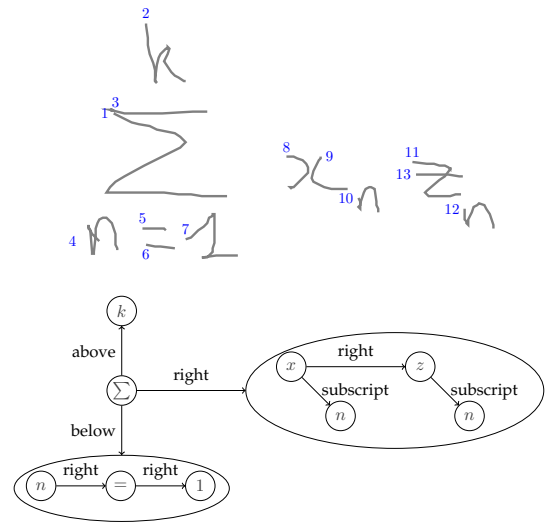

\IEEEPARstart{R}{ecognition} of online handwriting aims at finding the
best interpretation of a sequence of input
strokes~\cite{Plamondon:2000}. Roughly speaking,
handwriting data can be divided into two broad categories: text and
graphics. In text notation, symbols are usually composed  of strokes
that are consecutive relative to a time or spatial order; and symbols
themselves are also arranged according  to a specific order, for
example, from left to right. The ordering of symbols defines a single
adjacency, or relation type, between consecutive symbols. By contrast,
graphics encompass a variety of object types such as mathematical or
chemical expressions, diagrams, and tables. Symbols in graphics
notation are often composed of strokes that are not consecutive with
respect to neither time nor spatial order. Furthermore, a diversified
set of relations is possible between arbitrary pairs of symbols. See
Figure~\ref{fig:me_input}, for instance, where a handwritten
mathematical expression illustrates some characteristics of graphics
notation.

Due to the linear arrangement of symbols, text recognition can be
modeled as a parsing of one-dimensional (1D) data. On the other hand,
graphics are intrinsically two-dimensional (2D) data, requiring a
structural analysis, and there are no standard parsing methods as in
the 1D case. Parsing depends on symbol segmentation (or, stroke
grouping), symbol identification, and analysis of structural
relationship among constituent elements. Stroke grouping 
in texts is relatively simpler than in graphics as already mentioned.
Identification of segmented symbols include challenges such as
the possibly large number of symbol
classes, shape similarity between symbols in distinct classes, and
shape variability within a same class (e.g. arrows in flowcharts might
include arbitrary curves, and be directed towards any orientation).
Structural analysis involves the identification of relations between
symbols and a coherent integrated interpretation. The large variety of
relations might define complex hierarchical structures that increments the
difficulty in terms of efficiency and accuracy. There is a strong
dependency among the three tasks since symbol segmentation and
classification algorithms must often rely on structural or contextual
information to solve ambiguities, and structural analysis algorithms
depend on symbol identification to build coherent structures.

Although recognition of 2D objects is a subject of study since
  long ago~\cite{Marriott1998}, many of the efforts are still focused
  on solving
  specific aspects of the recognition process (e.g., detection of
  constituent parts or classification of components and their
  relations). A large number of works that tackle the
  entire recognition problem is clearly emerging, but they are often
  restricted to specific application domains and have
  limitations~\cite{Lin:2006,Chen:2015,Alvaro:2016}.

Motivated by the problem of online handwritten mathematical
expression recognition, we have examined issues related to the
recognition process and identified three features that are
desirable. The first feature is \textit{multilevel information
integration}. By multilevel information integration we mean
integrating symbol and structural level information to find the best
interpretation of a set of strokes. In mathematical expression
recognition, methods that seek information integration have already
been the concern of several
works~\cite{Awal:2010a,Alvaro:2011,MacLean:2013}, but it is still one
of the most challenging problems. The second feature is related to
\textit{model generalization}. Existing methods often limit the type
of expressions to be recognized (for instance, do not include
matrices), consider a fixed notation (for instance, it adopts either
$\sum_{i=1}^{n} x_i$ or $\displaystyle \sum_{i=1}^{n} x_i$), or limit
the set of mathematical symbols to be recognized. Any extensions
regarding these limitations may require major changes in the
recognition algorithms. The third feature is \textit{computational
  complexity management}. A general model often results in exponential
time algorithms, making its application unfeasible. Existing models
handle time complexity issues by adopting constraints that limit the
recognizable structures~\cite{MacLean:2013,Celik:2011}.

To deal with the issues described above, we have elaborated a general
framework for the recognition of online handwritten mathematical
expressions and then show its generality by building a flowchart recognition system 
using the same framework. We model a mathematical expression as a graph, and
represent the recognition problem as a graph parsing problem. The
recognition process is divided into three stages: (1) hypotheses
identification, (2) graph parsing, and (3) optimal interpretation
retrieval. The first stage computes a graph, called hypotheses graph,
that encodes plausible symbol interpretations and relations between
pairs of such symbols. The second stage parses the set of strokes to
find all interpretations that are valid according to a pre-defined graph
grammar, using the hypotheses graph to constrain the search space. The
parsing method is based on a recursive search of isomorphisms between
a labeled graph defined in the graph grammar and the ones derived from
the hypotheses graph. The last stage retrieves the most likely
interpretation based on a cost function that models symbol
segmentation, classification and structural information jointly.

Conceptually, the valid structures are defined through a graph
grammar and likely structures in the input stroke set are captured in
the hypotheses graph. Thus, the proposed framework enhances
independence of the parsing step with respect to specificities of the
mathematical notation considered. As a consequence, we have a flexible
framework with respect to different mathematical notations. For
instance, new expression structures can be included in the family of
expressions to be recognized by just including the structures in the
grammar rules. Similarly, the class of mathematical symbols to be
recognized can be extended by just including new symbol labels in the
grammar and in the hypotheses graph building procedure.

With respect to graphics in general, among them there is large
difference in the set of symbols and relations between symbols. Thus,
recognition techniques are often developed for a specific family of
graphics, introducing constraints that not only limit their
effectiveness, but also their adaptation to recognize different
families of graphics.
In spite of these differences, graphic notations share common concepts
-- a set of interrelated
symbols spread over a bidimensional space, organized in hierarchical
structures that are decisive to the interpretation.
We argue that the flexibility of the proposed framework encompasses
other families of graphics. This argument is supported by the fact
that graphs has already proven adequate to model graphics in
general. In addition, there are examples that show that families of
graphics can be specified by means of a graph
grammar~\cite{Rekers:1997,Han:2005,Alvaro:2016}. Moreover, hypotheses
graphs can be built based on data-driven approaches.

The main contributions of this work are thus twofold. First, we
present a general framework in which the parsing process is
independent of the family of graphics to be recognized and 
a control of the computational time is possible by means of a
hypotheses graph. Second, we demonstrate an effective application of
the framework to the recognition of mathematical expressions and
flowcharts.

The remaining of this text is organized as follows. In
Section~\ref{sec:related} we review some methods and concerns in
previous works related to the recognition of mathematical expression
and flowcharts, as these types of graphics served as the ground for
the development of the method described in this manuscript. We also
briefly comment on some works that proposed graph grammars for the
recognition of 2D data and influenced our work. In
Section~\ref{sec:framework} we detail the proposed framework. Then in
Section~\ref{sec:applications} we describe how the elements and
parameters required by the framework have been defined for the
recognition of mathematical expressions and flowcharts. In
Section~\ref{sec:experimentation} we present and discuss the
experimental results for both applications, and 
in Section~\ref{sec:conclusions} the conclusions and future works.

\section{Related work}
\label{sec:related}

In this section, we review some characteristics of the recognition
process in previous works, with emphasis on methods for
mathematical expression~\cite{Blostein:1997,Chan:2000,Zanibbi:2012}
and flowchart recognition~\cite{Miyao:2012,Carton:2013,Bresler:2014}.


Early works related to the recognition of mathematical
expressions were predominantly based on a sequential recognition
process consisting of the symbol segmentation, symbol identification
and structural analysis steps~\cite{Matsakis:1999,Tapia:2004,Zanibbi:2002}. 
However, a weakness of sequential methods is the fact that errors in
early steps are propagated to subsequent steps.
For instance, it might be difficult to determine if two handwritten
strokes with shape ``)" and ``(", close to each other, form a single
symbol ``x", or are the opening and the closing parentheses,
respectively. To solve this type of ambiguity, it may be necessary to
examine relations of the strokes with other nearby symbols or even
with respect to the global structure of the whole expression. 
This type of observation has motivated more recent works to consider
methods that integrate symbol and structural level interpretations
into a single process. Most of them are based on parsing methods as
described below.


Given an input stroke set, the goal of parsing is to find a parse tree
that ``explains" the structure of the stroke set, relative to a
predefined grammar. From a high-level perspective, parsing-based
techniques avoid sequential processing by generating several symbol
and relation interpretations, combining them to form multiple
interpretations of the whole input stroke set, and selecting the best
one according to a score (based on the whole structure).

An important element in parsing based approaches is the grammar. A
grammar defines how we model a (graphics) \textit{language}. For
mathematical expressions, most
approaches~\cite{Alvaro:2013,Awal:2009,Awal:2012,Yamamoto:2006,Simistira:2015}
use modifications of
context-free string grammars in Chomsky Normal Form\footnote{In a CNF,
all production rules either have the form $A \rightarrow a$, or $A
\rightarrow BC$, where $a$ is a terminal  and $A$, $B$, and $C$ are
non-terminals} (CNF). Such grammars define production rules of the
form $A \stackrel{r}{\rightarrow} BC$, where $r$ indicates a relation
between adjacent elements of the right hand side (RHS) of the
rule. For instance, expression $4^2$ can be modeled through a rule $TERM
\stackrel{superscript}{\rightarrow} NUMBER\ NUMBER$. However, as such
grammars impose the restriction of having at most two elements on the
RHS of a rule, structures with more than two components, like $2+4$,
or $\sum \limits_{i}^{n} x_i$,  must be modeled as a recursive
composition of pairs of components.
MacLean \textit{et. al.}~\cite{MacLean:2013} proposed 
\textit{fuzzy relational context free grammars} to overcome 
this limitation. They included production rules of the form: 
$A \stackrel{r}{\rightarrow} A_1 A_2 \ldots A_k$, where $r$ indicates a 
relation between adjacent elements of the RHS of the rule. 
However, the model assumes that the relation can only be 
of vertical or horizontal types.
\emph{Celik and Yanikoglu}~\cite{Celik:2011} use graph grammars
 with production rules of the form $A \rightarrow B$, where both $A$ and 
$B$ are graphs, and $B$ represents the components of a subexpression as
 vertices and their relations as edges. 
Graph grammar models offer more powerful representativeness compared to 
string grammars. However, the authors limit the 
grammars to have specific structures (each graph in a rule is
either a single vertex graph, or a \textit{star} graph -- a graph with
a single central vertex and surrounding vertices that are connected
only to the  central one), largely restricting the set of recognizable
expressions.

With respect to parsing, most algorithms proposed in the literature
for mathematical expressions are based on the CYK
algorithm~\cite{Younger:1967}. The CYK algorithm assumes that the input (in our
case) strokes form a sequence and the grammar is in CNF. Those based
on bottom-up approaches build a parse tree by first identifying
symbols (leaves) from single or groups of consecutive strokes, and
then combining the symbols recursively to form subcomponents
(subtrees), until obtaining a component that covers the whole input set.
To adapt the CYK algorithm to the recognition of mathematical expressions,
\emph{Yamamoto et. al.}~\cite{Yamamoto:2006} introduced an 
ordering of the strokes based on the input time. 
Other approaches avoid the stroke ordering assumption, 
but introduce different constraints to satisfy 
the decomposition of the input into pairs of
components~\cite{Alvaro:2013,Awal:2009,Awal:2012,Simistira:2015}. MacLean
\textit{et. al.}~\cite{MacLean:2013} proposed a top-down parsing
algorithm that does not assume grammars in CNF, but assumes 
that the input follows either a vertical or horizontal ordering (the  
\textit{fuzzy relational context free grammars} mentioned above).
Methods that use the CYK algorithm or others borrowed from the context
of string grammars must decompose the 2D input into a set of 1D
inputs. As there is no guarantee that such decomposition is possible,
these methods may present strong limitations with respect to parsable
2D structures and be completely inappropriate for
other types of 2D data.

On the other hand, methods that consider graph
grammars face computational complexity issues. A key step of any
parsing algorithm is the definition of how a stroke
set can be partitioned according to the RHS of a rule. Let us
consider a set of $n$ strokes. Assuming stroke ordering and a
CYK-based algorithm as
in~\cite{Yamamoto:2006,Alvaro:2013,Awal:2009,Awal:2012,Simistira:2015},
rules have at most two components in the RHS and therefore the number
of meaningful partitions is $O(n)$ -- we can
assign the first $i$ strokes to the first component and the rest for
the second, with $i \in \{1, \ldots, n-1\}$. On the other hand, if we do
not impose CNF, but keep the stroke ordering assumption as
in~\cite{MacLean:2013}, then a rule may have $k$ symbols on its RHS,
and the number of meaningful partitions is $O\binom{n}{k}$,
corresponding to $k-1$ split points on the sequence of $n$ strokes.
In graph grammars, without any restriction and a rule with $k$
vertices in the RHS, the number of partitions is $O(n^k)$ -- any
non-empty stroke subset can be mapped to any vertex.
Restricting the graph
structures in the grammar, for instance to star graph structures as done by
\emph{Celik and Yanikoglu}~\cite{Celik:2011}, is a way to manage
the parsing complexity. Note, however, that in this case the set of
recognizable expressions is constrained not only by the parsing
algorithm but also by the grammar.

\begin{figure}[htbp]
\centering
\includegraphics[width=0.9\linewidth]{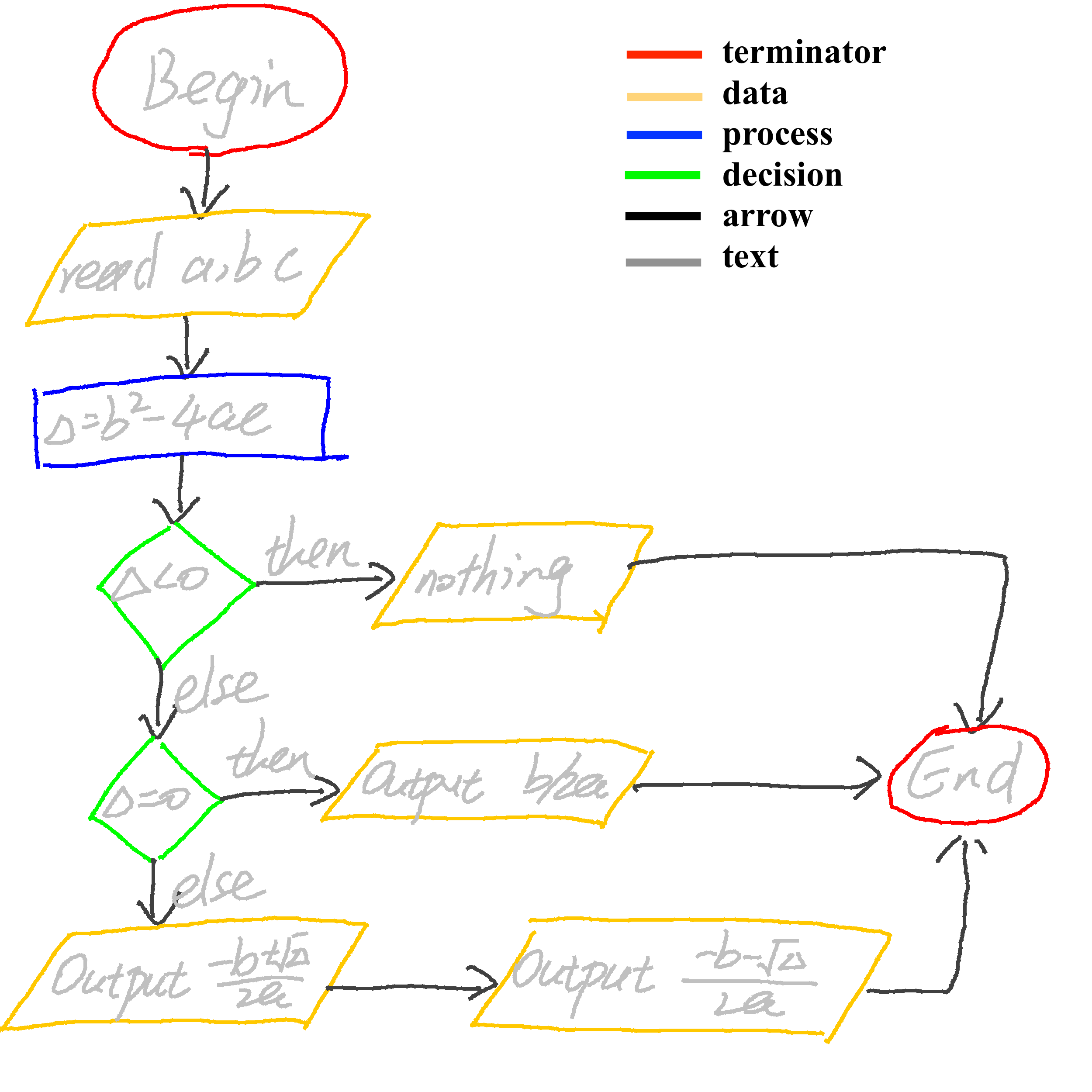}
\caption{Flowchart example. Strokes are colored according 
to the symbol type they belong to.}
\label{fig:fc_input}
\end{figure}

Flowcharts in general have a smaller symbol set than mathematical
expressions. However, their structure presents higher variance. For
instance, the flowchart in Figure~\ref{fig:fc_input} includes two
loops, and adjacent symbols can be located at any (vertical,
horizontal, or diagonal) position relative to each other, regardless
the relation type. In contrast, in mathematical expressions, for a
given relation type between two symbols (e.g. superscript) it is
expected that one symbol is located at some specific area
relative to the other (e.g. top-right). Thus, for flowcharts it may be
difficult to establish a spatial ordering of the input strokes.

To cope with the structural variance of diagrams, some approaches
introduce strong constraints in the input, as requiring all symbols to
have only one stroke~\cite{Yuan:2008}, or loop-like symbols to be
written by consecutive strokes~\cite{Miyao:2012}. With respect to
symbol recognition, detection of texts (or text box) and arrow symbols
are regarded as more difficult, as they do not present a fixed
shape. For instance, Carton~\textit{et. al.}~\cite{Carton:2013}
determine box symbols (like~\textit{decision}, and \textit{data}
structure) 
and then select the best interpretations using a deformation
metric. Text symbols are recognized only after box
symbols. Bresler~\textit{et. al.}~\cite{Bresler:2014} also first
recognize possible box and arrow symbols, and leave text recognition
as a last step. After symbol candidates are identified, the best
symbol combination is selected through a max-sum optimization process.

An interesting example of graph grammar use is described
in~\cite{Han:2005}. The authors propose an attributed graph grammar
that allow attributes to be passed from node to node in the grammar,
both vertically and horizontally, to describe a scene of man-made
objects. Projection of rectangles are used as primitives. However,
passage of attributes must be evaluated during parsing, making the
parsing algorithm be context-dependent. In~\cite{Rekers:1997}
entity-relationship diagrams are modeled by a context-sensitive graph
grammar with the ``left-hand side of every production being
lexicographically smaller than its right-hand side''. A critical part
of the parsing algorithm is to find matchings of the right-hand side
of a rule to replace the left-hand-side, making it very complex.
 
The above review on some characteristics related to the recognition of
2D data illustrates that existing methods present several restrictions
and limitations and clearly can not be easily transposed to the
recognition of other families of graphics.

In the method proposed in this work, instead a CYK-based algorithm
(that assumes a grammar in CNF), we define a graph grammar and use a
top-down parsing algorithm, similar to the one of~\cite{MacLean:2013},
but without assuming any ordering of the input strokes. To avoid
context-aware algorithms during parsing, we consider stroke partitions
drawn from a previously built hypotheses graph (see
Section~\ref{sec:parsing}) to match the right-hand side of the
rules. By doing this, we decouple the parsing algorithm from the
particularities of the family of graphics,
and achieve independence of the target notation. In addition, it is
important to note that target domain knowledge can be fully exploited
in the graph grammar definition and hypotheses graph building. This
characteristic makes the proposed method general enough to be applied
to the recognition of a variety of graphic notations.

%

\section{The proposed recognition framework}
\label{sec:framework}

The proposed recognition framework is composed of three main parts:
(1) \textbf{hypotheses graph generation}, (2) \textbf{graph parsing},
and (3) \textbf{optimal tree extraction}. In the first part, stroke
groups that are likely to represent symbols, and a set of possible
relations between these stroke groups are identified and stored as a
graph, called hypotheses graph. In the second part, valid 
interpretations (potentially multiple of them) are built from the
hypotheses graph by parsing it
according to a graph grammar. The interpretations found are stored in
a parse forest. Then, in the third part an optimal tree is extracted
from the parse forest, based on a scoring function. 

We first discuss the two main input data of the framework, a
handwritten input graphic to be recognized (a set of strokes) and a
graph grammar, and then detail the three parts, keeping an abstraction
level suitable for the recognition of a variety of graphics in
general. Concepts are illustrated using mathematical expressions as
examples. Implementation related details regarding the application of
the framework to the recognition of mathematical expressions and
flowcharts are presented in Section~\ref{sec:applications}.

\subsection{Stroke set}
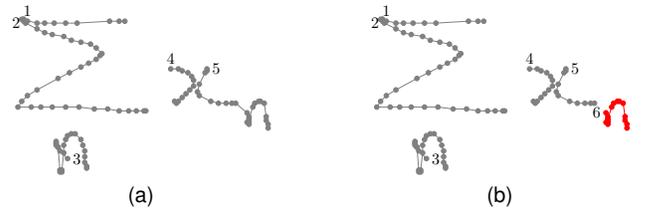
\begin{figure}[htb]
\centering
 \ \ \ \ \subfloat[\label{fig:concepts.a}]{\resizebox{0.4\linewidth}{!}{\begin{tikzpicture}[x=1, y=-1,scale=1.2]

\draw[gray] plot[mark=*, mark size = 1.2] coordinates {(238, 135) (236, 134)(236, 135)(236, 134)(235, 134) (243, 136) (247, 136) (252, 136) (258, 136) (264, 136) (281, 135) (286, 135) (289, 135)};

\draw[gray] plot[mark=*, mark size = 1.2] coordinates {(237, 136) (243, 139) (247, 141) (251, 142) (257, 143) (261, 145) (266, 146) (271, 147) (274, 149) (277, 152) (276, 153) (274, 155) (270, 157) (266, 159) (260, 162) (254, 165) (243, 171) (238, 174) (235, 177) (233, 180) (240, 180) (245, 180) (251, 180) (258, 180) (265, 180) (273, 181) (280, 181) (286, 182) (291, 182) (295, 182) (299, 182) (300, 182)};

\draw[gray] plot[mark=*, mark size = 1.2] coordinates {(259, 207) (255, 203) (254, 201) (253, 199) (253, 198) (253, 199) (253, 200) (254, 201) (255, 213) (255, 214) (256, 214) (256, 213) (257, 204) (257, 200) (258, 197) (259, 195) (261, 194) (263, 194) (266, 197) (267, 200) (268, 203) (268, 206) (268, 209) (269, 211) (269, 212)};

\draw[gray] plot[mark=*, mark size = 1.2] coordinates {(313, 160) (317, 160) (319, 161) (322, 162) (324, 164) (325, 166) (325, 169) (323, 171) (321, 173) (319, 175) (317, 177) (315, 177) (316, 178)};

\draw[gray] plot[mark=*, mark size = 1.2] coordinates {(332, 160) (332, 161) (331, 162) (328, 168) (327, 172) (328, 174) (334, 177) (338, 178) (342, 178) (345, 178) (347, 178) (353, 183) (354, 185) (354, 187) (353, 188) (354, 189) (354, 188) (354, 186) (356, 180) (357, 178) (359, 177) (360, 177) (362, 178) (363, 185) (363, 187) (364, 189) (364, 191)};

\draw (238, 135) node[anchor=south] {$1$};

\draw (237, 136) node[anchor=east] {$2$};

\draw (259, 207) node[anchor=west] {$3$};

\draw (313, 160) node[anchor=south] {$4$};

\draw (332, 160) node[anchor=west] {$5$};


\end{tikzpicture}}}
 \hfill
\subfloat[\label{fig:concepts.b}]{\resizebox{0.4\linewidth}{!}{\begin{tikzpicture}[x=1, y=-1,scale=1.2]

\draw[gray] plot[mark=*, mark size = 1.2] coordinates {(238, 135) (236, 134)(236, 135)(236, 134)(235, 134) (243, 136) (247, 136) (252, 136) (258, 136) (264, 136) (281, 135) (286, 135) (289, 135)};

\draw[gray] plot[mark=*, mark size = 1.2] coordinates {(237, 136) (243, 139) (247, 141) (251, 142) (257, 143) (261, 145) (266, 146) (271, 147) (274, 149) (277, 152) (276, 153) (274, 155) (270, 157) (266, 159) (260, 162) (254, 165) (243, 171) (238, 174) (235, 177) (233, 180) (240, 180) (245, 180) (251, 180) (258, 180) (265, 180) (273, 181) (280, 181) (286, 182) (291, 182) (295, 182) (299, 182) (300, 182)};

\draw[gray] plot[mark=*, mark size = 1.2] coordinates {(259, 207) (255, 203) (254, 201) (253, 199) (253, 198) (253, 199) (253, 200) (254, 201) (255, 213) (255, 214) (256, 214) (256, 213) (257, 204) (257, 200) (258, 197) (259, 195) (261, 194) (263, 194) (266, 197) (267, 200) (268, 203) (268, 206) (268, 209) (269, 211) (269, 212)};

\draw[gray] plot[mark=*, mark size = 1.2] coordinates {(313, 160) (317, 160) (319, 161) (322, 162) (324, 164) (325, 166) (325, 169) (323, 171) (321, 173) (319, 175) (317, 177) (315, 177) (316, 178)};

\draw[gray] plot[mark=*, mark size = 1.2] coordinates {(332, 160) (332, 161) (331, 162) (328, 168) (327, 172) (328, 174) (334, 177) (338, 178) (342, 178) (345, 178) (347, 178) };

\draw[red] plot[mark=*, mark size = 1.2] coordinates {(353, 183) (354, 185) (354, 187) (353, 188) (354, 189) (354, 188) (354, 186) (356, 180) (357, 178) (359, 177) (360, 177) (362, 178) (363, 185) (363, 187) (364, 189) (364, 191)};

\draw (238, 135) node[anchor=south] {$1$};

\draw (237, 136) node[anchor=east] {$2$};

\draw (259, 207) node[anchor=west] {$3$};

\draw (313, 160) node[anchor=south] {$4$};

\draw (332, 160) node[anchor=west] {$5$};

\draw (353, 183) node[anchor=east] {$6$};


\end{tikzpicture}}}
\ \ \ \ \\
\caption{Handwritten expressions representing 
$\sum \limits_{n}^{} x_n$. Each expression is composed 
of a set of strokes, where each stroke is a sequence 
of bidimensional coordinates (dots in gray). In (a), stroke $5$ belongs 
to two symbols. In (b), each stroke belongs to only one symbol.}
\label{fig:concepts}
\end{figure}

Online handwriting consists of a set of strokes. Each stroke is,
typically, a sequence of point coordinates sampled from the moment a
writing device (such as a stylus) touches the screen up to the moment it
is released. We assume that each stroke belongs to only one
symbol (this assumption is common when dealing with  
handwritten graphics). Otherwise, a preprocessing step could be applied to
split a stroke that is part of two or more symbols. These concepts are
illustrated in Figures~\ref{fig:concepts.a} and~\ref{fig:concepts.b}.

\subsection{Graph grammar model}
\label{sec:graph_grammar}

A graph grammar~\cite{Pflatz:1969} defines a \textit{language} of
graphs. We denote a graph $G$ as a pair $(V_G, E_G)$, where
$V_G$ represents the set of vertices of $G$ and $E_G$ represents the set
of edges of $G$. A labeled graph is a graph with labels in its vertices
and edges. Hereafter we assume labeled graphs, with labels defined by
a function $l$ that assigns symbol labels (in a set $SL$) to vertices
and relation labels (in a set $RL$) to edges.
We define a family of graph grammars, called~\emph{Graphic grammars},
to model graphics as labeled graphs.
 
\begin{definition} A ~\emph{graphic grammar} is a tuple 
$M=(N, T, I, R)$ where:
\begin{itemize}
 \item $N$ is a set of non-terminal nodes (or non-terminals);
 \item $T$ is a set of terminal nodes (or terminals), such that
 $N \cap T = \emptyset$ (for convenience we denote elements in $T$
 using the same names used for the labels in $SL$); 
 \item $I$ is a non-terminal, called initial node;
 \item $R$ is a set of production (or rewriting) rules of the form
 $A := B$ where $A$ is a non-terminal node and $B=(V_B,E_B)$ is a
   connected graph with 
   label $l(v) \in N \cup T$ for each  $v \in V_B$, and label $l(e)
   \in RL$ for each $e \in E_B$.
\end{itemize}
\end{definition}

Note that $M$ is a context-free graph grammar~\cite{Pflatz:1969}.
The language defined by a graphic grammar $M=(N, T, I, R)$ is a
(possibly infinite) set of connected labeled graphs and is denoted
$\mathcal{L}(M)$. Similarly to string grammars, a labeled graph $G$
belongs to $\mathcal{L}(M)$ if $G$ can be derived (or generated) from
the initial non-terminal node $I$ by successively applying production
rules in $R$, until obtaining a graph with only terminal nodes.

\begin{figure}[htbp]
\centering
\includegraphics[width=\linewidth]{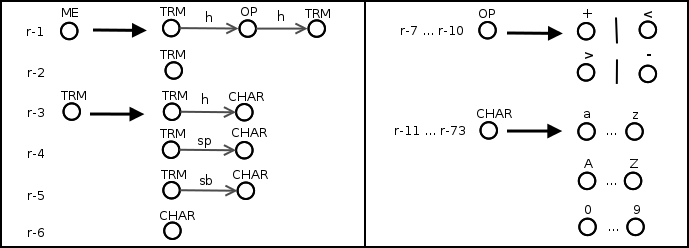}
\caption{Graph grammar that models basic mathematical expressions. The
  grammar is defined by non-terminals $N=\{ME, TRM, OP, CHAR\}$,
  relation labels $RL = \{sp, sb, h\}$, terminals $T =\{+,-, <,
  >, a, \ldots,z, A, \ldots, Z, 0, \ldots, 9\}$, rules $R = \{r-1,
  \ldots, r-73\}$, and $ME$ at the left hand side graph of rule
  $r-1$ is the initial node. Abbreviations: $ME$ = mathematical
  expression, $sp$ = superscript, $sb$ = subscript, $h$ = horizontal, $TRM$ =
  term, $OP$ = operator, $CHAR$ = character.}
\label{fig:grammar}
\end{figure}

Figure~\ref{fig:grammar} shows a graphic grammar that models
simple arithmetic and logical expressions. Each production rule
defines the replacement of a non-terminal, a single vertex graph $G_l$
at the left hand side (LHS) of the rule, with a graph $G_r$ at the
right hand side (RHS).

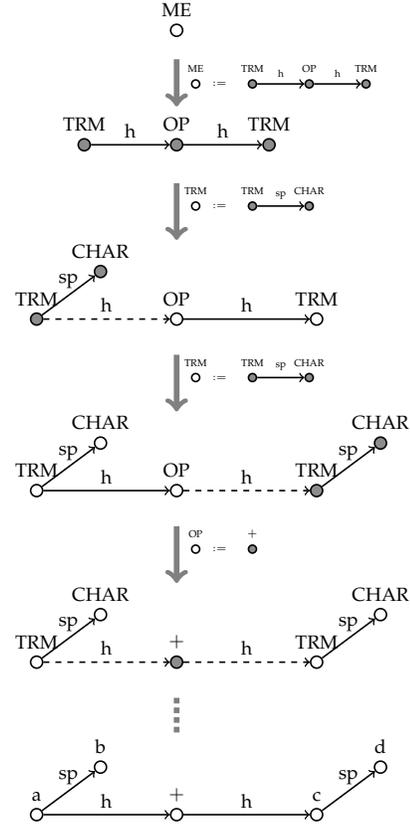
\begin{figure}[htbp]
\centering
\resizebox{0.65\linewidth}{!}{
%
%

\begin{tikzpicture}[->]
  \def\ggDifference{-3em}
  \def\nodetoNodeDistance{4.5em}
  \def\nodetoNodeDoubleDistance{7em}
  \def\nodetoNodeXDominatedDistance{3em}
  \def\nodetoNodeYDominatedDistance{2.5em}
  \def\baseNodeYDistance{4ex}
  \def\supHeight{6em}
\tikzstyle{vertex}=[ellipse,fill=white,minimum size=6pt,inner sep=0pt, draw=black, thick]
  \tikzstyle{boldVertex}=[ellipse,fill=black!45,minimum size=6pt,inner sep=0pt, draw=black, thick]
  \tikzstyle{littlevertex}=[ellipse,fill=white,minimum size=4pt,inner sep=0pt, draw=black, thick]
  \tikzstyle{littleboldVertex}=[ellipse,fill=black!45,minimum size=4pt,inner sep=0pt, draw=black, thick]
   \tikzstyle{grammarGraph}=[rectangle, fill=white,minimum height=4em, minimum width=16em, inner sep=0pt, draw=white]
   \tikzstyle{grammarGraph1n}=[rectangle, fill=white,minimum height=3em, minimum width=4em, inner sep=0pt, draw=white]
   \tikzstyle{grammarGraph1Supn}=[rectangle, fill=white,minimum height=\supHeight, minimum width=20em, inner sep=0pt, draw=white]
   \tikzstyle{grammarGraph2Supn}=[rectangle, fill=white,minimum height=\supHeight, minimum width=20em, inner sep=0pt, draw=white]

\tikzstyle{myarrows}=[line width=1mm,draw=gray,-triangle 45,postaction={draw, line width=3mm, shorten >=4mm, -}]

  \node[grammarGraph1n] (gg1) at (5,0) {};
  \node[vertex, label = ME] (N-ME) at (gg1.center) {};
  
  \node[minimum height=1em, minimum width=2em, inner sep=0pt, draw=white] (graphL) at ([yshift = -2.3em, xshift = 1em] gg1.center) {};
   \node[littlevertex, label = \tiny ME] (gl) at (graphL.south) {};
   \node[minimum height=1em, minimum width=2em, inner sep=0pt, draw=white]  (graphR) [right of = graphL] {};
   \node[littleboldVertex, label = \tiny TRM] (gr1-1) at (graphR.south) {};
   \node[littleboldVertex, label = \tiny OP] (gr1-2) [right of = gr1-1] {};
   \node[littleboldVertex, label = \tiny  TRM] (gr1-3) [right of = gr1-2] {};
   \draw[->, thick] (gr1-1) -- (gr1-2) node [above, midway] {\tiny h};
   \draw[->, thick] (gr1-2) -- (gr1-3) node [above, midway] {\tiny h};
   \node at ([xshift = 1em] gl.east) {\tiny $:=$};
  
   \node[grammarGraph] (gg2) at ([yshift=\ggDifference - 1.5em]gg1.south) {};
  \node[boldVertex, label = OP] (N-OP1) at (gg2.center) {};
  \node[boldVertex, label = TRM] (N-TRM1) at ([xshift=-\nodetoNodeDistance]N-OP1.west) {};
    \node[boldVertex, label = TRM] (N-TRM2) at ([xshift=\nodetoNodeDistance]N-OP1.east) {};
    \draw [->, thick](N-OP1) -- (N-TRM2) node[above,midway] {h};
    \draw [->, thick] (N-TRM1) -- (N-OP1) node[above,midway] {h};
   
\node[minimum height=1em, minimum width=2em, inner sep=0pt, draw=white] (graphL2) at ([yshift = -2.7em, xshift = 1em] gg2.center) {};
   \node[littlevertex, label = \tiny TRM] (vl2) at (graphL2.south) {};
   \node[minimum height=1em, minimum width=2em, inner sep=0pt, draw=white]  (graphR2) [right of = graphL2] {};
   \node[littleboldVertex, label = \tiny TRM] (gr2-1) at (graphR2.south) {};
   \node[littleboldVertex, label = \tiny CHAR] (gr2-2)  [right of = gr2-1] {};
   \draw[->, thick] (gr2-1) -- (gr2-2) node [above, midway] {\tiny sp};
   \node at ([xshift = 1em] vl2.east) {\tiny $:=$};
    
    \node[grammarGraph1Supn] (gg3) at ([yshift=\ggDifference - \supHeight /2]gg2.south) {};
    \node[vertex, label = OP] (N-OP3) at ([yshift=\baseNodeYDistance]gg3.south) {};
    \node[boldVertex, label = TRM] (N-TRM3) at ([xshift=-\nodetoNodeDoubleDistance]N-OP3.west) {};
    \node[boldVertex, label = CHAR] (N-CHAR3) at ([xshift=\nodetoNodeXDominatedDistance, yshift=\nodetoNodeYDominatedDistance]N-TRM3.east) {};
    \node[vertex, label = TRM] (N-TRM3-2) at ([xshift=\nodetoNodeDoubleDistance]N-OP3.east) {};
    \draw [->, thick](N-OP3) -- (N-TRM3-2) node[above,midway] {h};
    \draw [->, dashed, thick] (N-TRM3) -- (N-OP3) node[above,midway] {h};
    \draw [->, thick] (N-TRM3) -- (N-CHAR3) node[above,midway] {sp};
    
    \node[minimum height=1em, minimum width=2em, inner sep=0pt, draw=white] (graphL3) at ([yshift = -3.7em, xshift = 1em] gg3.center) {};
   \node[littlevertex, label = \tiny TRM] (vl3) at (graphL3.south) {};
   \node[minimum height=1em, minimum width=2em, inner sep=0pt, draw=white]  (graphR3) [right of = graphL3] {};
   \node[littleboldVertex, label = \tiny TRM] (gr3-1) at (graphR3.south) {};
   \node[littleboldVertex, label = \tiny CHAR] (gr3-2) [right of = gr3-1] {};
   \draw[->, thick] (gr3-1) -- (gr3-2) node [above, midway] {\tiny sp};
   \node at ([xshift = 1em] vl3.east) {\tiny $:=$};
    
    \node[grammarGraph2Supn] (gg4) at ([yshift=\ggDifference - \supHeight /2]gg3.south) {};
    \node[vertex, label = OP] (N-OP4) at ([yshift=\baseNodeYDistance]gg4.south) {};
    \node[vertex, label = TRM] (N-TRM4) at ([xshift=-\nodetoNodeDoubleDistance]N-OP4.west) {};
     \node[vertex, label = CHAR] (N-CHAR4-1) at ([xshift=\nodetoNodeXDominatedDistance, yshift=\nodetoNodeYDominatedDistance]N-TRM4.east) {};
     \node[boldVertex, label = TRM] (N-TRM4-2) at ([xshift=\nodetoNodeDoubleDistance]N-OP4.east) {};
    \node[boldVertex, label = CHAR] (N-CHAR4-2) at ([xshift=\nodetoNodeXDominatedDistance, yshift=\nodetoNodeYDominatedDistance]N-TRM4-2.east) {};
    
    \draw [->, dashed, thick](N-OP4) -- (N-TRM4-2) node[above,midway] {h};
    \draw [->, thick] (N-TRM4) -- (N-OP4) node[above,midway] {h};
    \draw [->, thick] (N-TRM4) -- (N-CHAR4-1) node[above,midway] {sp};
    \draw [->, thick] (N-TRM4-2) -- (N-CHAR4-2) node[above,midway] {sp};
    
    \node[minimum height=1em, minimum width=2em, inner sep=0pt, draw=white] (graphL4) at ([yshift = -3.7em, xshift = 1em] gg4.center) {};
   \node[littlevertex, label = \tiny OP] (vl4) at (graphL4.south) {};
   \node[minimum height=1em, minimum width=2em, inner sep=0pt, draw=white]  (graphR4) [right of = graphL4] {};
   \node[littleboldVertex, label = \tiny $+$] (gr4-1) at (graphR4.south) {};
   \node at ([xshift = 1em] vl4.east) {\tiny $:=$};
    
    \node[grammarGraph2Supn] (gg5) at ([yshift=\ggDifference - \supHeight /2]gg4.south) {};
    \node[boldVertex, label = $+$] (N-OP5) at ([yshift=\baseNodeYDistance]gg5.south) {};
    \node[vertex, label = TRM] (N-TRM5) at ([xshift=-\nodetoNodeDoubleDistance]N-OP5.west) {};
     \node[vertex, label = CHAR] (N-CHAR5-1) at ([xshift=\nodetoNodeXDominatedDistance, yshift=\nodetoNodeYDominatedDistance]N-TRM5.east) {};
     \node[vertex, label = TRM] (N-TRM5-2) at ([xshift=\nodetoNodeDoubleDistance]N-OP5.east) {};
    \node[vertex, label = CHAR] (N-CHAR5-2) at ([xshift=\nodetoNodeXDominatedDistance, yshift=\nodetoNodeYDominatedDistance]N-TRM5-2.east) {};
    
    \draw [->, dashed, thick](N-OP5) -- (N-TRM5-2) node[above,midway] {h};
    \draw [->, dashed, thick] (N-TRM5) -- (N-OP5) node[above,midway] {h};
    \draw [->, thick] (N-TRM5) -- (N-CHAR5-1) node[above,midway] {sp};
    \draw [->, thick] (N-TRM5-2) -- (N-CHAR5-2) node[above,midway] {sp};

    \node[grammarGraph2Supn] (gg6) at ([yshift=\ggDifference - \supHeight /2 + 1em]gg5.south) {};
    \node[vertex, label = $+$] (N-OP6) at ([yshift=\baseNodeYDistance]gg6.south) {};
    \node[vertex, label = a] (N-TRM6) at ([xshift=-\nodetoNodeDoubleDistance]N-OP6.west) {};
     \node[vertex, label = b] (N-CHAR6-1) at ([xshift=\nodetoNodeXDominatedDistance, yshift=\nodetoNodeYDominatedDistance]N-TRM6.east) {};
     \node[vertex, label = c] (N-TRM6-2) at ([xshift=\nodetoNodeDoubleDistance]N-OP6.east) {};
    \node[vertex, label = d] (N-CHAR6-2) at ([xshift=\nodetoNodeXDominatedDistance, yshift=\nodetoNodeYDominatedDistance]N-TRM6-2.east) {};
    
    \draw [->, thick](N-OP6) -- (N-TRM6-2) node[above,midway] {h};
    \draw [->, thick] (N-TRM6) -- (N-OP6) node[above,midway] {h};
    \draw [->, thick] (N-TRM6) -- (N-CHAR6-1) node[above,midway] {sp};
    \draw [->, thick] (N-TRM6-2) -- (N-CHAR6-2) node[above,midway] {sp};

     \draw [->, line width = 1mm, gray] (gg1) -- (gg2); 
     \draw [->, line width = 1mm, gray] (gg2) -- (gg3); 
     \draw [->, line width = 1mm, gray ] (gg3) -- (gg4);
     \draw [->, line width = 1mm, gray ] (gg4) -- (gg5); 
     \draw [dotted, -, line width = 1mm, gray ] (gg5) -- (gg6.north);

\end{tikzpicture}

} 
\caption{Generation of a graph that represents the expression 
$a^b + c^d$. At each 
rule application, the replacing graph nodes are depicted in 
dark gray. Edges that link the replacing graph with 
the host graph are depicted with dashed arrows.
Rule applications after the fourth one are not shown.}
\label{fig:graphGeneration}
\end{figure}

Figure~\ref{fig:graphGeneration} shows a graph generation process
using the grammar of Figure~\ref{fig:grammar}. Rules are applied
sequentially, starting with non-terminal $ME$, until all elements in the
generated graph are terminals. Dashed arrows correspond to edges that
link the replacing graphs with the host graph.

\begin{figure}[htb]
\centering
\resizebox{0.8\linewidth}{!}{
%
%

\begin{tikzpicture}[->]
  \def\ggDistance{9em}
  \def\nodetoNodeDistance{4.5em}
  \def\nodetoNodeDoubleDistance{7em}
  \def\nodetoNodeXDominatedDistance{3em}
  \def\nodetoNodeYDominatedDistance{2.5em}
  \def\baseNodeYDistance{4ex}
  \def\supHeight{6em}
  \tikzstyle{vertex}=[ellipse,fill=white,minimum size=6pt,inner sep=0pt, draw=black, thick]
  \tikzstyle{boldVertex}=[ellipse,fill=black!45,minimum size=6pt,inner sep=0pt, draw=black, thick]
   \tikzstyle{grammarGraph}=[rectangle, fill=white,minimum height=10em, minimum width=9em, inner sep=0pt, draw=white]
   \tikzstyle{grammarGraph1n}=[rectangle, fill=white,minimum height=3em, minimum width=4em, inner sep=0pt, draw=white]
   \tikzstyle{grammarGraph1Supn}=[rectangle, fill=white,minimum height=\supHeight, minimum width=21.5em, inner sep=0pt, draw=white]
   \tikzstyle{grammarGraph2Supn}=[rectangle, fill=white,minimum height=\supHeight, minimum width=26em, inner sep=0pt, draw=white]
  
  
  \node[grammarGraph] (gg1) at (5,0) {};
  \node[vertex,label=C] (N-C1) at (gg1.north) {};
  \node[vertex,label=B] (N-B1) at ([xshift=-\nodetoNodeDistance, yshift=-\nodetoNodeDistance]N-C1.center) {};
  \node[vertex,label=A] (N-A1) at ([xshift=\nodetoNodeDistance, yshift=-\nodetoNodeDistance]N-C1.center) {};
  \draw [->, thick](N-A1) -- (N-B1) node[above,midway] {x};
  \draw [->, thick](N-B1) -- (N-C1) node[above,midway] {y};
  \draw [->, thick](N-C1) -- (N-A1) node[above,midway] {z};
 \draw [-,very thick, black!70] (gg1.center) -- (gg1.south) {};
  
  
  \node[grammarGraph] (gg2) at ([yshift=-\ggDistance, xshift=-\ggDistance] gg1.south) {};
  \node[vertex,label=C] (N-C2) at (gg2.center) {};
  \node[vertex,label=B] (N-B2) at ([xshift=-\nodetoNodeDistance, yshift=-\nodetoNodeDistance]N-C2.center) {};
  \node[boldVertex,label=D] (N-A2) at ([xshift=\nodetoNodeDistance, yshift=-\nodetoNodeDistance]N-C2.center) {};
  \node[boldVertex,label=E] (N-C2-2) at ([xshift=\nodetoNodeDistance]N-A2.center) {};
  \draw [->, thick, dashed](N-A2) -- (N-B2) node[above,midway] {x};
  \draw [->, thick](N-B2) -- (N-C2) node[above,midway] {y};
  \draw [->, thick, dashed](N-C2) -- (N-A2) node[above,midway] {z};
  \draw [->, thick](N-A2) -- (N-C2-2) node[above,midway] {w};
  
  \node[grammarGraph] (gg3) at ([yshift=-\ggDistance, xshift=\ggDistance] gg1.south) {};
  \node[vertex,label=C] (N-C3) at (gg3.center) {};
  \node[vertex,label=B] (N-B3) at ([xshift=-\nodetoNodeDistance, yshift=-\nodetoNodeDistance]N-C3.center) {};
  \node[boldVertex,label=E] (N-A3) at ([xshift=\nodetoNodeDistance, yshift=-\nodetoNodeDistance]N-C3.center) {};
  \node[boldVertex,label=D] (N-C3-2) at ([xshift=\nodetoNodeDistance]N-A3.center) {};
  \draw [->, thick, dashed](N-A3) -- (N-B3) node[above,midway] {x};
  \draw [->, thick](N-B3) -- (N-C3) node[above,midway] {y};
  \draw [->, thick, dashed](N-C3) -- (N-A3) node[above,midway] {z};
  \draw [->, thick] (N-C3-2) -- (N-A3) node[above,midway] {w};
  
  \draw[->,very thick, black!70] (gg1.south) -- (gg2.east) node[left, midway, align = center] {$\epsilon(r) = \{(V, D) | (V, A) \in G\} \cup $ \\ 
   $\{(D, V) | (A,V) \in G\}$};
   \draw[->,very thick, black!70] (gg1.south) -- (gg3.west) node[right, midway, align = center] {$\epsilon(r) = \{(V, E) | (V, A) \in G\} \cup $ \\ 
   $\{(E, V) | (A,V) \in G\}$};

   \node[minimum height=1em, minimum width=2em, inner sep=0pt, draw=white] (graphL) at ([yshift = -2em, xshift = 2em] gg1.center) {};
   \node[vertex, label = A] (A) at (graphL.south) {};
   \node[minimum height=1em, minimum width=2em, inner sep=0pt, draw=white]  (graphR) [right of = graphL] {};
   \node[boldVertex, label = D] (D) at (graphR.south) {};
   \node[boldVertex, label = E] (E) [right of = D] {};
   \draw[->, thick] (D) -- (E) node [above, midway] {w};
   \node at ([xshift = 1em] A.east) {$:=$};
   \node at ([xshift = -1em] A.west) {r:};

\end{tikzpicture}
%
}
\caption{Graph transformation with two different embeddings. 
The top graph is transformed through rule $r$. Each embedding 
defines edges between vertices that are linked to vertex $A$ 
of the top graph with vertex $D$ (left hand side embedding) or 
$E$ (right hand side embedding) of the replacing graph. 
Dashed arrows represent the edges defined by each embedding.}
\label{fig:embedding}
\end{figure}
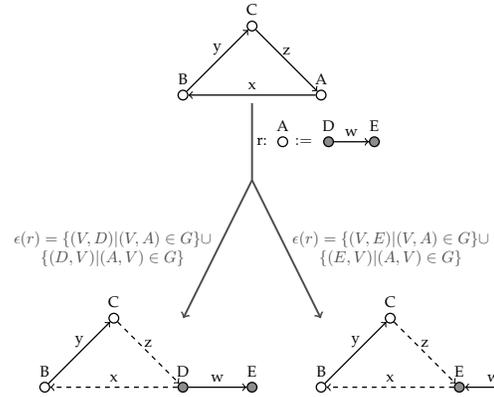

The definition of how a replacing graph should be linked to a host
graph $G$ is called \textit{embedding}~\cite{Pflatz:1969}, and it should
be specified for each production rule. Formally, given a production
rule $G_l := G_r$, its application consists in replacing a subgraph
$G_l$ of $G$ with $G_r$ and the embedding defines how $G_r$ will be
attached to $G \setminus G_l$. The \textit{attachment} may be defined
by a set of edges that link the replacing graph $G_r$ to $G \setminus
G_l$. For instance, Figure~\ref{fig:embedding} shows two different embeddings
for a same production rule, and the graphs generated for each
embedding.

The embedding specification depends on the desired language. It is
possible to define a same embedding specification for all rules, as we
do for mathematical expressions (see
Section~\ref{sec:applications}). An embedding can
also take spatial information into consideration, for example by
including edges only between spatially close vertices. More detailed
examples of embeddings are provided in Section~\ref{sec:applications},
through applications to the recognition of mathematical expressions and
flowcharts. To ensure that the generated graphs are connected, we
assume that every embedding is specified in such a way that its
application generates connected graphs.

\subsection{Hypotheses graph generation}
\label{sec:hypotheses:graph}

\begin{figure}[htb]
\centering
\includegraphics[width=0.85\linewidth]{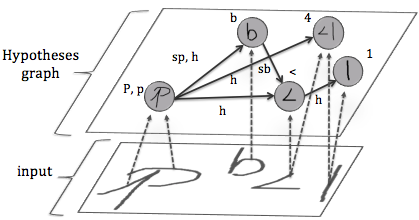}
\caption{Hypotheses graph example. Vertices represent symbol 
hypotheses and edges represent relations between symbols. The
labels associated to symbols and relations indicate their 
most likely interpretations.}
\label{fig:hypotheses_graph}
\end{figure}

Given a set of strokes $S$, we define a hypotheses graph as an
attributed graph $H=(V_H, E_H)$, where $V_H$ is a set of symbol
hypotheses and $E_H$ is a set of relation hypotheses computed from
$S$. Each symbol hypothesis $v \in V_H$ corresponds to a subset of
$S$, denoted as $stk(v)$, and has as an attribute a list $L(v) =
\{(l_i,s_i), i=1,\ldots, k_v \}$ of likely interpretations. Each of
these interpretations
$(l_i,s_i)$ consists of a symbol label $l_i \in SL$ and its respective
likelihood score $s_i \in [0,1]$. Note that a stroke may be shared by
multiple symbol hypotheses. Relation hypotheses (edges in $E_H$) are
defined over pairs of disjoint symbol hypotheses (i.e., hypotheses
such that their stroke sets are disjoint), and also have as an
attribute a list of likely relation interpretations denoted
$L(e)$. Relation labels are in $RL$. Figure~\ref{fig:hypotheses_graph}
shows a handwritten mathematical expression and a hypotheses graph
calculated from it.

To build a hypotheses graph, machine learning methods are effective in
identifying groups of strokes that may form symbols and, similarly,
relations among them (see application example in Section~\ref{sec:applications}).
Since many stroke groups do not correspond to an actual symbol and
many pairs of symbols are not directly related each other within a
graphic, rather than training classifiers to identify only true
hypotheses, those that do not represent any symbol or relation can be
included as elements of an additional class,
called \textit{junk}. Training data can be extracted from within the
graphic, together with surrounding context, in order to improve
rejection of false hypotheses. As will become clear later, hypotheses
graphs play an important role to constrain the search space during the
parsing process. A high precision and recall in the identification of
symbol hypotheses and relations is thus desirable to efficiently
constrain the search space.

\subsubsection{Label list pruning}
To define the labels and respective likelihood scores of symbol and
relation hypotheses, we could use the confidence scores returned by
the respective classifiers. However, to manage complexity, only class
labels that present high confidence scores should be kept. Selecting
the labels to be kept based on a fixed global confidence threshold
value is not adequate since label distributions vary greatly among
symbols and relations. An effective method to select the most
likely labels for each hypothesis $h$ is described next.

Let $\{ (l_i,s_i), i=1,..., n_h\}$ be the pairs of labels and
respective scores initially attributed to $h$, sorted in descending
order according to the likelihood scores $s_i$. Then, given a
distribution threshold $tr$ (between 0 an 1), we define the minimum
number of $k$ top ranked labels whose confidences sum up to at
least $tr$:
\begin{equation}
\label{eq:labels_selection_intro}
k = \arg \min\limits_{x} \sum\limits_{i=1}^{x} s_i > tr
\end{equation}
Hypothesis $h$ is rejected if it presents highest score for the $junk$
class label and if that score is above the threshold $tr$. Otherwise,
we set $L(h) = \{ (l_i,s_i) : i=1,\ldots,k\}$.
We define label pruning thresholds $t_{\mathrm{symb}}$ for symbols and
$t_{\mathrm{rel}}$ for relations.

\subsection{Graph parsing}
\label{sec:parsing}

The goal of the parsing process is to build a parsing tree that
explains the set of input strokes $S_{\mathrm{input}}$, according to a
grammar. Since there
might be more than one interpretation, multiple trees might be
generated, possibly sharing subtrees each other. Thus, they will be
stored in a parse forest. 

\begin{figure}[htbp]
\centering
\includegraphics[width=0.7\linewidth]{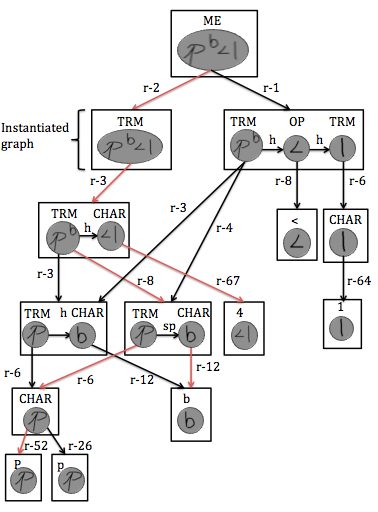}
\caption{A parse forest representing multiple interpretations of 
a mathematical expression. Labels on arrows refer to grammar rules of 
Figure~\ref{fig:grammar}. 
Red arrows represent a parse tree that corresponds to 
the interpretation ``$P^b4$''.}
\label{fig:parseForest}
\end{figure}

Figure~\ref{fig:parseForest} shows a parse
forest calculated from the hypotheses graph of
Figure~\ref{fig:hypotheses_graph}, using the graph grammar of
Figure~\ref{fig:grammar}. 
As can be seen, the root node (top of the figure) corresponds to the
starting non-terminal $ME$. Two branches are generated from rules
associated to $ME$. The left branch is generated by applying rule 2 and
the right branch by applying rule
1. Note that, for each rule, any of the resulting partition of the
strokes induces a graph that is isomorphic to the RHS graph of the
respective rule. The same principle holds for the remaining of the
nodes.

The parsing process follows a top-down approach. To understand the
parsing process, a key step is to understand how a stroke set is
partitioned when a rule is applied. More specifically, given a set of
strokes $S$ and a non-terminal $NT$, for each rule $A:=B$
associated to $NT$, we must find every partition of $S$ that is a
valid matching to $B$. A partition of $S$ is a matching to $B$ if
its number of parts is equal to the number of vertices of $B$, so that
each part can be assigned to one vertex in $B$. A matching is valid if
the following two conditions hold: (1) the partition of $S$ induces a
graph that is isomorphic to $B$, and (2) each subset of strokes
assigned to a vertex of $B$ must be parsable according to the grammar.

Supposing the number of vertices in $B$ is $k$ and the number of
strokes in $S$ is $n$, without any constraint, the total number of
possible stroke partitions to be examined to generate the valid
matchings would
be $O(k^n)$. Exhaustively examining each of these partitions is not
computationally practical.

A main strategy of our method is to constrain the number of partitions
to be examined with the aid of the hypotheses graph. We assume that
all meaningful interpretations are present in the hypotheses graph as
a subgraph. Thus, before starting the parsing process, we build the
set of all stroke groups, denoted hereafter as $STK$, underlying any
valid connected subgraph of $H$. Note that these stroke groups must not
contain repeated strokes, i.e., a valid subgraph is one in which a
same stroke is not present twice. Furthermore, not all stroke groups
will be necessarily parsable. The relation between two stroke groups
is also recorded in STK as being the same between the corresponding
subgraphs. Hence, during parsing, the search space of valid matchings
will be restricted to those present in $STK$. Once a valid matching is
found, an instance of $B$, which we call \emph{instantiated graph},
will be recursively parsed and will become a \emph{parsed graph} when
each of its vertices is successfully parsed.

The complete algorithm is described next. For the sake of
simplification, we will assume that the input grammar contains only
two types of rules: \textit{terminal}
and \textit{non-terminal}. Terminal rules are productions of the form
$A := b$, where the RHS graph $b$ is a single vertex graph, with
labels in the terminal set, such as rules from r-7 to r-73 of the
grammar of Figure~\ref{fig:grammar}. Non-terminal rules are
productions of the form $A := B$, where $B$ is a graph
containing one or more vertices, each of them with non-terminal
labels, such as rules r-1 to r-6 of the grammar of
Figure~\ref{fig:grammar}. Thus, Algorithm 1 considers only these two
types of rules. Its extension to treat rules that contain both
terminals and non-terminals in its right-hand side is a
straightforward combination of the previous two cases.

\begin{figure}[htb]
\includegraphics[width=\linewidth, trim={4.5cm 14cm 4.5cm 4.2cm},clip]{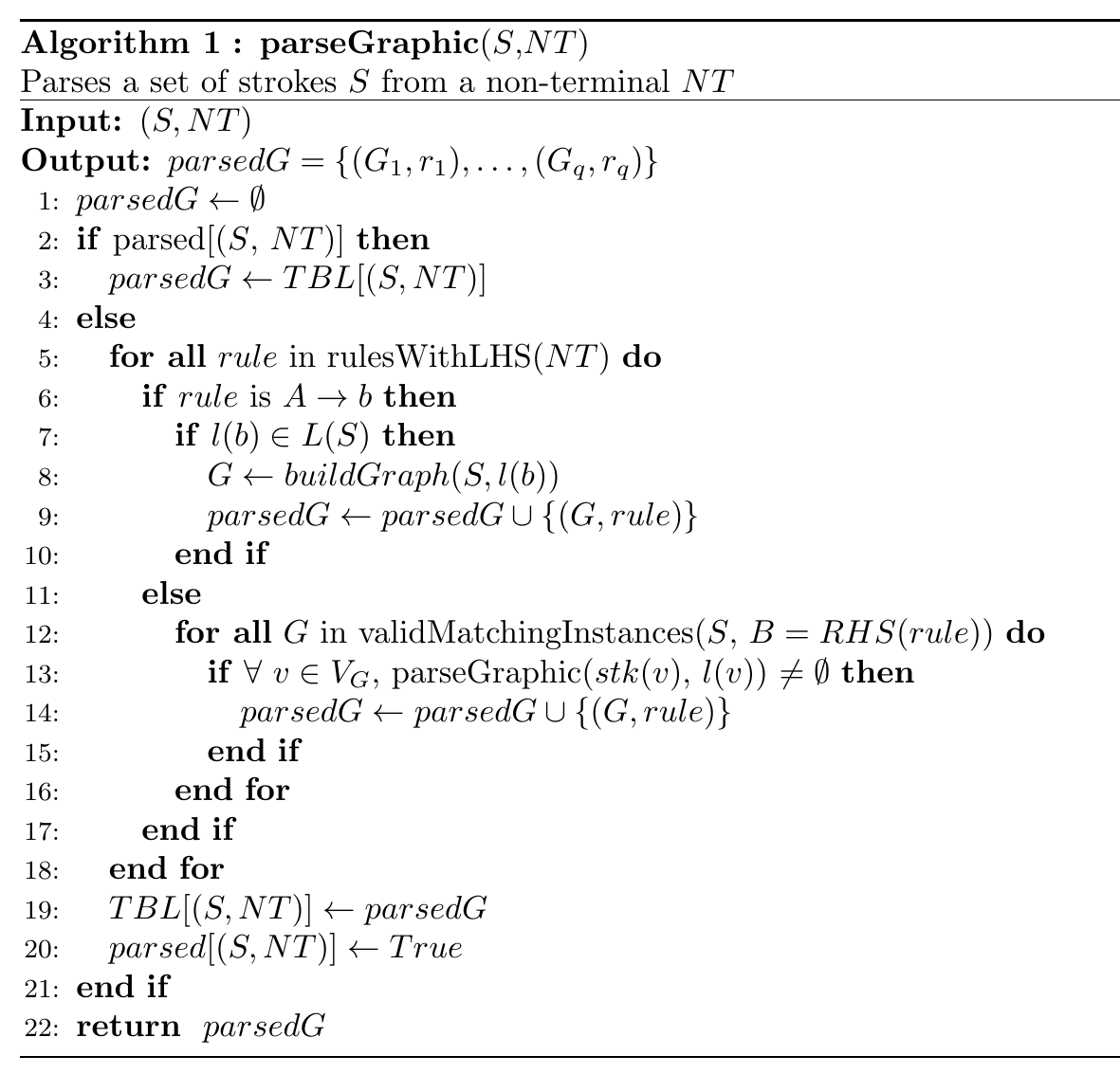}
\end{figure}

Algorithm 1 receives as inputs a stroke set $S = \{stk_1, \ldots,
stk_n\}$ and a non-terminal $NT$. Initially, the set of strokes is the
whole input set $S_{\mathrm{input}}$ and the non-terminal is the
starting node $I$. Then, it applies each of the production rules that
have $NT$ as the LHS graph and returns a set ($parsedG$) containing
all parsed graphs, together with the respective rules that
``generated'' them.

To avoid recomputation, a global table $TBL$ indexed by pairs $(S
= \{stk_1, \ldots, stk_n\}, NT)$ is used. An entry in TBL is of the form
$TBL[(S, NT)] = \{(G_1, r_1), \ldots, (G_q,r_q)\}$ where $G_i$ is a
parsed graph and $r_i$ is the rule that ``generated'' $G_i$. 
At the end of the algorithm, if the
pair $(S, NT)$ is not parsable, the corresponding entry in $TBL$ is
empty.



Lines 2-3 verify if the pair $(S,NT)$ has already been processed. If
so, results are retrieved from $TBL$ and returned. Otherwise, lines 5-18
iterate over the rules that have $NT$ in its LHS graph. If the rule is
of terminal type (lines 6-10), it suffices to check if the RHS vertex
label, $l(b)$, is contained in the set of labels $L(S)$ attributed
to the underlying stroke set. This verification is done by checking if
the stroke set $S$ corresponds to a vertex in the hypotheses graph 
and if the label set of the corresponding vertex includes $l(b)$. Then
a single vertex graph is built and stored together with the
rule in $parsedG$. If the rule is of non-terminal type (lines 11-17),
for each valid matching between $S$ and $B$ (line 12) we verify if the
instantiated graph is parsable.
The parsing result, either a list of parsed graphs, or an empty list
(in case of parsing failure), is added to $TBL$. As already mentioned,
table $TBL$ is used to avoid parsing recomputation of pairs
$(S,NT)$. At the end of the parsing process, the parse forest can be
extracted from $TBL$ by traversing it starting from index
$(S_{\mathrm{input}},I)$.

\subsubsection{Pruning strategies}
\label{sec:pruning}

Besides constraining the partitions to be examined to only those
formed by stroke groups that underlie a subgraph of $H$, there are other
strategies that can be used to speed up computation. For example, 
determining the maximum and minimum size of non-terminal nodes is a
strategy that has been previously used in text parsing~\cite{Grune:2008}. The sizes,
in terms of graphic symbols or strokes, can be computed directly from the
grammar. Based on these numbers, during parsing any stroke subsets
that are out of the min-max ranges do not need to be evaluated. This
information can be calculated when building $STK$. Moreover, to find
valid matching partitions, the minimum and maximum sizes of the stroke
subsets already matched to some vertices can be used to determine the
minimum and maximum size of the stroke groups that still can be
matched to the rest of the nodes.

Another useful information is to explore the knowledge that a
non-terminal can generate only a specific subgroup of the
terminals. For instance, in the grammar of Figure~\ref{fig:grammar},
non-terminal $OP$ can generate only symbols $+$, $-$, $<$, or
$>$. Thus, stroke subsets that do not contain any hypothesis with one of such
labels as terminals are not evaluated during the parsing
process. Analogously, stroke groups that correspond to symbol and
relation hypotheses with high mean junk score can be
disregarded. Specifically, stroke subsets with a certain number
(five, for example) symbol hypotheses, having mean junk score,
including both symbol and relation labels, above a given junk
threshold $t_{\mathrm{junk}}$ will not be considered. This pruning is
mainly useful when the symbol and relation hypotheses have a large
number of labels. High mean junk score indicates that it is unlikely
that the underlying group of strokes is parsable.


\subsection{Optimal parse tree extraction}

Once a parse forest is built, the final step consists in traversing
it to extract the best tree (interpretation). 
To characterize what is an optimal tree (best interpretation), we
first define a cost function for trees. Roughly stating, an
interpretation will be considered of low cost if its corresponding
parse tree includes substructures with high confidence scores.

We introduce a few notations that will be helpful. Let $x$ denote a
node in the parse forest. Let $G_x=(V_x,E_x)$ be the graph
instantiated at node $x$. Each vertex $v \in V_x$ has an underlying
set of strokes, $stk(v)$. For each terminal vertex $v \in V_x$ there
will be a pair $(label(v),score(v)) \in SL\times[0,1]$ and for each
edge $e \in E_x$ will be a pair $(label(e),score(e)) \in
RL\times[0,1]$.

The cost of a tree can be computed bottom-up. We first define
individual costs relative to symbols and relations, and then define
how to combine the two to determine the cost of a tree.
Let $t$ be a parse tree and let $x$ be a node in $t$. Let $child(x)$
denote the child nodes of $x$. The subtree in $t$ with root at $x$ is
denoted $t_x$. We first assign to a node $x$
a symbol cost $J_s(x)$:
\begin{equation}
J_s(x) = \left\{
\begin{array}{ll}
-\log\,score(v), & \mbox{if $x$ is terminal,}\\
          & \mbox{with $V_x =\{v\}$,}\\
\displaystyle \sum_{y \in child(x)}J_s(y) & \mbox{if $x$ is non-terminal,}
\end{array}
\right.
\end{equation}
and a relation cost $J_r(x)$:
\begin{equation}
J_r(x) = \sum_{e \in E_x} -\log\,score(e) + \sum_{y \in child(x)}J_r(y)
\end{equation}


Then, the cost of $t_x$ is defined as
\begin{equation}
\label{eq:tree_cost}
J(t_x) = \frac{\alpha}{n_s} J_s(x) + \frac{1-\alpha}{n_r} J_r(x)
\end{equation}
where $n_s$ and $n_r$ are, respectively, the number of symbols and
relations under $t_x$. Parameter $\alpha$ weights both types of costs,
and could be adjusted to give more relevance to one or to the other.

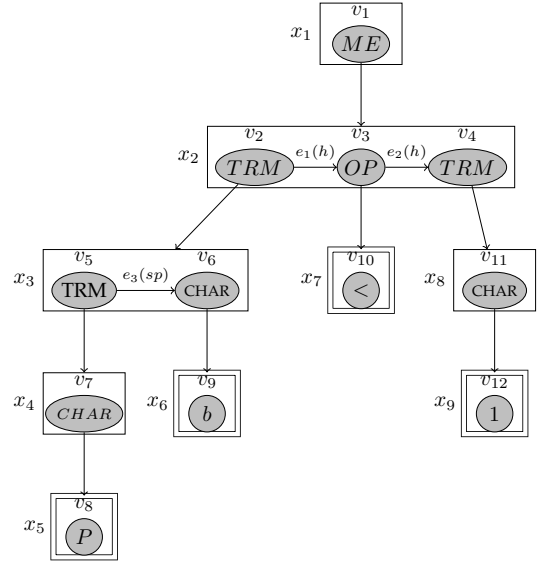
\begin{figure}[htb]
\centering
\resizebox{0.8\linewidth}{!}{
%
%
%
\begin{tikzpicture}[node distance=6em]
\def \yshiftFromCenter{-0.5em}
\def \nodetoNodeDistance{5.2em}
\def\ggDifference{-3em}
\def \ggXDifference{5em}
\tikzstyle{vertex}=[ellipse,fill=black!25,minimum size=17pt,inner sep=0pt, draw=black]
\tikzstyle{grammarGraph1n}=[rectangle, fill=white, minimum height=3em, minimum width=4em, inner sep=0pt, draw=black]
\tikzstyle{grammarGraph1terminal}=[rectangle, fill=white,double=white,double distance =2pt, minimum height=3em, minimum width=3em, inner sep=0pt, draw=black]
\tikzstyle{grammarGraph2n}=[rectangle, fill=white, minimum height=3em, minimum width=10em, inner sep=0pt, draw=black]
\tikzstyle{grammarGraph1Supn}=[rectangle, fill=white,minimum height=\supHeight, minimum width=21.5em, inner sep=0pt, draw=black]
   \tikzstyle{grammarGraph2Supn}=[rectangle, fill=white,minimum height=\supHeight, minimum width=26em, inner sep=0pt, draw=black]
    \tikzstyle{grammarGraph3n}=[rectangle, fill=white,minimum height=3em, minimum width=15em, inner sep=0pt, draw=black]
   
\node[grammarGraph1n] [label={left:$x_1$}] (x1){};
\node[vertex] [label=$v_1$] (v1) at ([yshift  = \yshiftFromCenter]x1.center){$ME$};

\node[grammarGraph3n] [label={left:$x_2$}] (x2) at ([yshift=\ggDifference - 1.5em]x1.south)  {};
\node[vertex] (v3) [label=$v_3$] at ([yshift  = \yshiftFromCenter]x2.center){$OP$};
\node[vertex] (v2) [label=$v_2$] at ([xshift  = -\nodetoNodeDistance]v3.center){$TRM$};
\node[vertex] (v4) [label=$v_4$] at ([xshift  = \nodetoNodeDistance]v3.center){$TRM$};

\node[grammarGraph1terminal] [label={left:$x_7$}] (x7) at ([yshift=\ggDifference - 1.5em]x2.south)  {};
\node[vertex] (v10) [label=$v_{10}$] at ([yshift  = \yshiftFromCenter]x7.center){$<$};

\node[grammarGraph1n] [label={left:$x_8$}] (x8) at ([xshift = \ggXDifference]x7.east)  {};
\node[vertex] (v11) [label=$v_{11}$, font = \scriptsize] at ([yshift  = \yshiftFromCenter]x8.center){CHAR};

\node[grammarGraph1terminal] [label={left:$x_9$}] (x9) at ([yshift=\ggDifference - 1.5em]x8.south)  {};
\node[vertex] (v12) [label=$v_{12}$] at ([yshift  = \yshiftFromCenter]x9.center){$1$};

\node[grammarGraph2n] [label={left:$x_3$}] (x3) at ([xshift = -9em]x7.west)  {};
\node[vertex] (v5) [label=$v_{5}$] at ([xshift  = -3em, yshift  = \yshiftFromCenter]x3.center){TRM};
\node[vertex] (v6) [label=$v_{6}$, right of=v5, font = \scriptsize] {CHAR}; 

\node[grammarGraph1n] [label={left:$x_4$}] (x4) at ([xshift  = -3em, yshift=\ggDifference - 1.5em]x3.south)  {};
\node[vertex] (v7) [label=$v_{7}$, font = \scriptsize] at ([yshift  = \yshiftFromCenter]x4.center){$CHAR$};

\node[grammarGraph1terminal] [label={left:$x_5$}] (x5) at ([yshift=\ggDifference - 1.5em]x4.south)  {};
\node[vertex] (v8) [label=$v_{8}$] at ([yshift  = \yshiftFromCenter]x5.center){$P$};

\node[grammarGraph1terminal] [label={left:$x_6$}] (x6) at ([xshift  = 3em, yshift=\ggDifference - 1.5em]x3.south)  {};
\node[vertex] (v9) [label=$v_{9}$] at ([yshift  = \yshiftFromCenter]x6.center){$b$};

\draw [->] (v1) -- (x2) node[above,midway] {};
\draw [->] (v2) -- (x3) node[above,midway] {};
\draw [->] (v5) -- (x4) node[above,midway] {};
\draw [->] (v7) -- (x5) node[above,midway] {};
\draw [->] (v6) -- (x6) node[above,midway] {};
\draw [->] (v3) -- (x7) node[above,midway] {};
\draw [->] (v4) -- (x8) node[above,midway] {};
\draw [->] (v11) -- (x9) node[above,midway] {};

\draw [->] (v2) -- (v3) node[above,midway, font=\scriptsize] {$e_1 (h)$};
\draw [->] (v3) -- (v4) node[above,midway, font=\scriptsize] {$e_2 (h)$};
\draw [->] (v5) -- (v6) node[above,midway, font=\scriptsize] {$e_3 (sp)$};

\end{tikzpicture}

}
\caption{Parse tree of expression $P^b < 1$, extracted from the 
parse forest of Figure~\ref{fig:parseForest}. Nodes are indexed 
as $x_i$, $i = 1,\ldots , 9$. Similarly, vertices and 
edges of the instantiated graphs are respectively indexed as $v_j$, 
for $j=1, \ldots , 12$, and $e_k$, for $k=1, \ldots , 3$. Nodes with terminal 
symbols are depicted with double line borders.}
\label{fig:treeCost}  
\end{figure}

An example of a tree is shown in Figure~\ref{fig:treeCost}. Its root
node is $x_1$ and thus the tree is denoted $t_{x_1}$. The cost of tree
$t_{x_1}$ is given in Eq.~\ref{eq:tree_cost_example}.

\begin{align}
\label{eq:tree_cost_example}
J(t_{x_1})  = & 
 \frac{\alpha}{4} \Big(J_s(v_8)  + J_s(v_9)  + J_s(v_{10})  + J_s(v_{12})\Big)  +  \nonumber \\ 
 &   \left( \frac{1- \alpha}{3} \right) \Big( J_r(e_1) + J_r(e_2) + J_r(e_3) \Big) 
\end{align}

In order to extract the best tree, the cost of each tree in the parse
forest must be computed. Since the trees in the parse forest share
subtrees, this fact can be explored to avoid computing the cost of a
shared subtree repeatedly. In addition, from an application point of
view, being able to efficiently retrieve a number of best parse trees
rather than just the best one is often desirable. We borrow ideas from
the tree extraction technique, in the context of string grammars,
proposed by \textit{Boullier et al.}~\cite{Boullier:2009}. Given a
parse forest, they proposed a method that builds a new parse forest
with a fixed number of $n$-best trees, using a bottom-up approach.
The resulting parse forest can be further processed to improve the
recognition result, for example, by doing a re-ranking of the trees, a
processing that could be too expensive to be done in the original
parse forest.

Note that there might be multiple subtrees with root at a node $x$ in
the parse forest. For instance, in the parse forest of
Fig~\ref{fig:parseForest}, the vertex in the bottom left non-terminal
node graph has two possible derivations (``P'' or ``p''). Whenever
there are multiple derivations from a non-terminal vertex, only one of
them will be present in a parse tree. Thus, given a node $x$ in the
parse forest, let us denote by $t_x^{(i)}$, $i\in I_x$, the spanned
trees from $x$. The number of possible trees in the forest is
combinatorial with respect to the multiple subtrees spanned from the
nodes in a path from the root node to a leaf node.

We use a bottom-up approach to compute, for each node $x$ in the
forest, a list of subtrees spanned from it. This information is
kept as a table in the node, and each row of the table stores
information to recover one of the spanned trees (specifically, it
stores the partition of the stroke set resulting from the
corresponding derivation). After the bottom-up process finishes,
individual trees can be extracted by performing a top-down traversal,
starting from each row of the table at the root node of the
forest. The best tree, according to the specified cost, is the one
recovered by starting the traversal from the first row of the table.

However, since there might be a large number of parse trees in the
forest, a naive application of the method described above may
be computationally prohibitive. To overcome this problem, a pruning
strategy can be applied during the bottom-up step to keep table sizes
manageable: for each table, spanned trees that have a cost much higher
than the best tree are discarded.
To compute relative differences of cost, let $minJ(x)$ be the minimum
cost tree spanned from $x$. Then, given $t_{\mathrm{pr}} \in [0,1]$, a
spanned tree $t_x^{(i)}$ is kept if
\begin{equation}
\label{eq:tree_pruning}
|J(t_x^{(i)}) - minJ(x)| < t_{\mathrm{pr}} * minJ(x)\,.
\end{equation}
This strategy resembles the one proposed in~\cite{Boullier:2009}, but
it differs in the sense that while they keep a fixed number of best
trees, we keep only the relatively likely ones. The more ambiguous the
input, the more parse trees are kept. The pruning threshold 
$t_{\mathrm{pr}}$ can be empirically estimated.

\section{Applications}
\label{sec:applications}

The application of the framework requires the definition of some key
elements. First, a graph grammar that models the family of graphics to
be recognized must be defined. A set of labels for the relations ($RL$) and
for the symbols ($SL$), including \emph{junk}, must be defined. Second,
a hypotheses graph generated from the set of input strokes, with
symbol labels in $SL$ and relation labels in $RL$, must be
provided. Terminal nodes of the grammar are named using the labels in
$SL$, while edges in the graphs of the grammar are labeled using
labels in $RL$. For parsing, an embedding method must be defined for
each grammar rule. In this section, we detail how these elements as
well as important parameter values have been defined for the
recognition of mathematical expressions and flowcharts. Results and
discussions are presented in the next section. The grammars in
\texttt{xml} format are available at{\small~\url{www.vision.ime.usp.br/~frank.aguilar/grammars/}}.

Before applying the recognition method itself, we applied to the set
of strokes the smoothing and resampling methods described
in~\cite{Delaye:2013}. Smoothing removes abrupt trajectory changes in the
strokes and resampling makes point distribution uniform -- equally
spaced -- along the strokes. In the evaluating  
datasets, each stroke belongs to only one symbol; thus no additional preprocessing 
was needed.

\subsection{Recognition of mathematical expressions}

\subsubsection{Dataset and Grammar}

We use the CROHME-2014 dataset~\cite{crohme:2014}. It consists of
handwritten expressions divided into training and test sets, with
$9,507$ and $986$ expressions, respectively.
The expressions include $101$ symbol classes, and six relation classes
(\textit{horizontal} as in ``$ab$'', \textit{above} as in ``$\sum
\limits_{}^{x}$'', \textit{below} as in ``$\sum \limits_{x}^{}$'',
\textit{superscript} as in ``$a^b$'', \textit{subscript} as in
``$a_b$'', and \textit{inside} as in ``$\sqrt{x}$''). CROHME-2014
dataset provides a string grammar for the corresponding \LaTeX\ expressions. Based on that
string grammar, we defined a graph grammar with $205$ production
rules, including the rules to generate the $101$ symbol labels
(terminals).

To define the embeddings, we use the concept of baseline. A baseline
in a graph is defined as a maximal path whose connecting edges have
only the \textit{horizontal (h)} label (this definition can be seen as
a graph version of the baseline definition of~\cite{Zanibbi:2002}). A
baseline is considered nested to a vertex
$v$ if it is connected to $v$ by an edge $(v,v')$, where $v'$ is the
first vertex of the baseline. A baseline that is nested to no vertex
is called dominant baseline. Note that a baseline may consist of a
single vertex.

Then, the embedding is defined as follows. Let $r: G_l\!:=\!G_r$ be a
rule and let $v'$ be the leftmost and $v''$ be the rightmost vertices
of the dominant baseline of $G_r$. Let also $G$ be a graph with an
occurrence of $G_l$, identified as a vertex $u \in V_G$.  The
embedding associated to the application of rule $r$ on $G$ replaces
$u$ with $G_r$, generating an updated graph $G'$, such that $V_{G'} =
V_{G}\setminus \{u\} \cup V_{G_r}$ and $E_{G'} = [ E_{G}\setminus
  (\{(u',u): u'\in V_G\} \cup \{(u,u'): u'\in V_G\}) ] \cup \epsilon$ where
\begin{equation}
\epsilon = \{(u',v'): (u',u)\in E_G\} \cup \{(v'',u') : (u,u')\in E_G\}.
\end{equation}
In other words, all edges that were incident on $u$ will be made
incident to $v'$ and all edges that were originated from $u$ will be
made originating from $v''$.


\subsubsection{Hypotheses graph building}

To generate the hypotheses graph, we used the symbol segmentation and
classification methods described in~\cite{Frank:2015,Frank:2014b}, along 
with the spatial relation classification methods described
in~\cite{Frank:2016}. They are based on multilayer neural networks
with shape context descriptor~\cite{Belongie:2002}, and images created
from symbols and relations, including neighboring strokes to be used as
contextual information. The networks use a softmax output which is
then converted to a cost measure (applying the negative logarithm to
the output) in order to be used in the cost function defined
in Eq.~\ref{eq:tree_cost}. 

An important parameter to build the hypotheses graph is the symbol and
relation label pruning thresholds, $t_{\mathrm{symb}}$ and
$t_{\mathrm{rel}}$ (see Eq.~\ref{eq:labels_selection_intro}). These
threshold values determine how many and which labels will be attached
to each vertex and edge. Since during the parsing process the
partitions of the stroke set and labels are constrained by the
hypotheses graph, the achievable maximum recognition rates are bounded
by possibilities encoded in the hypotheses graph.

From the training set, we randomly selected $950$ expressions (about  
$10\%$) to serve as a validation set and used the rest for training.
Using the trained symbol and relation classifiers, we evaluated the
effect of varying values of $t_{\mathrm{symb}}$ and $t_{\mathrm{rel}}$
on the validation set. For each threshold value we computed the
symbol, relation and complete expression recalls, that is, how
many of each of these components were present in the hypotheses graph.

\begin{figure}[h]
 \centering
\includegraphics[width=\linewidth]{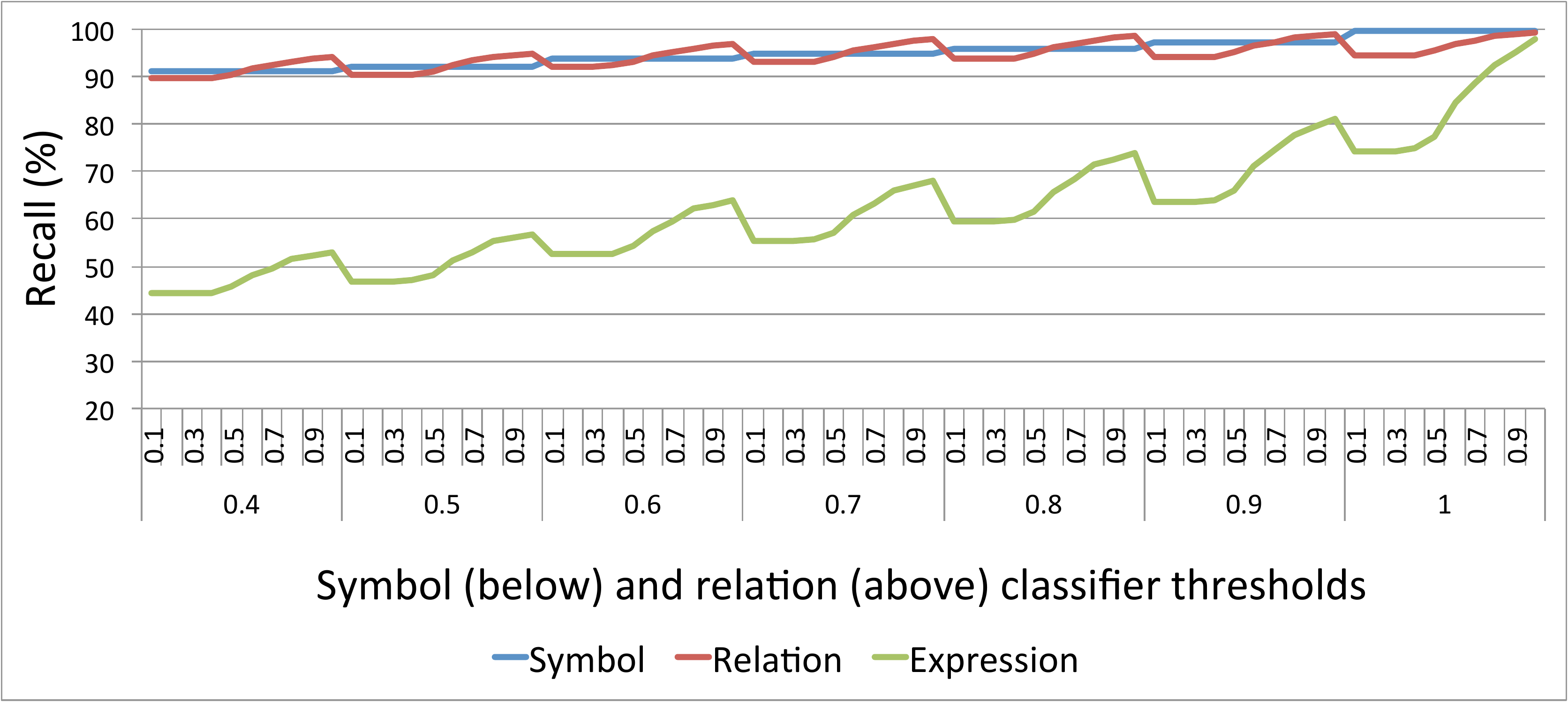}
 
\caption{Symbol, relation and expression level recall of the
  hypotheses graph generation step. For each symbol classification
  threshold $t_{\mathrm{symb}}$ in the range $[0.4-1.0]$, relation classification
  threshold $t_{\mathrm{rel}}$ is varied in the range $[0.1-1.0]$.}
\label{fig:recall_h_graph}
\end{figure}

Figure~\ref{fig:recall_h_graph} shows the results relative to this
evaluation, over $t_{\mathrm{symb}}$ in the range $[0.4-1]$ (for values less
than 0.4, the performance was similar to the case of 0.4) and
$t_{\mathrm{rel}}$ in the range $[0.1-1]$. Note that this evaluation is
concerned with verifying how many of the elements of interest are, in
fact, present in the hypotheses graph; it is not related with parsing.

We can see in Figure~\ref{fig:recall_h_graph} that even for the lowest
threshold values the recall of symbols and relation is about
$90\%$. For complete expressions (i.e. all symbols and relations of the expressions are in the hypothesis graph), however, the recall for the lowest
threshold values is $40\%$. If symbol classification threshold is set
to $1$, $99,75\%$ of the symbols are correctly included. Since in this
case no stroke group is rejected, $99,75\%$ is also the percentage of
symbols identified by the stroke grouping method. If, in addition,
we also set the relation classification threshold to $1$, almost all
relations and expressions are included ($99,45\%$ and $98.11\%$,
respectively).

\subsubsection{Graph parsing and tree extraction}

\begin{figure}[h]
 \centering
\includegraphics[width=\linewidth]{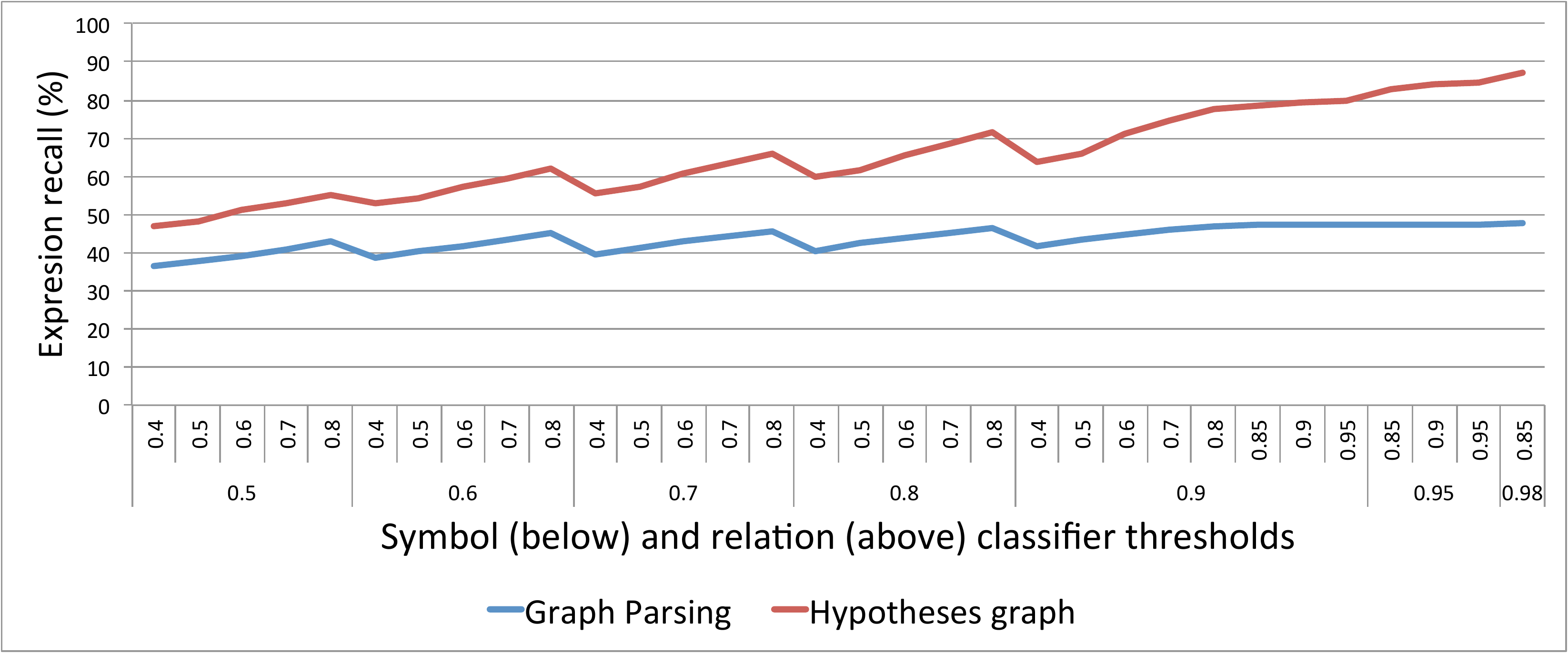}
 
\caption{Expression recall obtained at graph parsing and hypotheses 
graph generation steps, for different symbol and relation 
thresholds.}
\label{fig:graph_parsing_recall_validation}
\end{figure}

We also analyzed the effect of different values of $t_{\mathrm{symb}}$ and
$t_{\mathrm{rel}}$ on the recall after parsing. We set the maximum value for
$t_{\mathrm{symb}}$ to $0.98$ and for $t_{\mathrm{rel}}$ to $0.85$, as parsing large
expressions with thresholds larger than those takes much time to be
considered in a real application. In this evaluation, for optimal tree
extraction we set $\alpha = 0.5$ (same weight for the symbol and
relation costs, see Eq.~\ref{eq:tree_cost}) and $t_{\mathrm{pr}}=0.1$
(tree pruning threshold,
see Eq.~\ref{eq:tree_pruning}). Figure~\ref{fig:graph_parsing_recall_validation}
shows the expression recall obtained by the parsing method and the
corresponding recall obtained by the hypotheses graph generation step
(note that the second indicates the maximum achievable
recall). Although for values above $t_{\mathrm{symb}} = 0.9$ and
$t_{\mathrm{rel}} = 0.8$ no considerable improvements are observed in
the parsing recall, the gap between hypotheses graph recall and
parsing recall increases up to about $40\%$. Thus, we chose
$t_{\mathrm{symb}} = 0.98$ and $t_{\mathrm{rel}} = 0.85$, as these
values allow to keep more hypotheses and can be useful during
parsing of unseen expressions (better generalization).

Using $t_{\mathrm{symb}} = 0.98$ and $t_{\mathrm{rel}} = 0.85$, we
have also evaluated the effect of different values of $t_{\mathrm{pr}}$
(tree pruning threshold) and $\alpha$ (weighting in the cost function)
on tree extraction on validation set. Through this evaluation, we set
$t_{\mathrm{pr}}=0.1$ and $\alpha=0.4$ (this choice was based on the
best expression recall).

\subsection{Recognition of flowcharts}
\label{sec:flowchart_application}

\subsubsection{Dataset and grammar}

We use the flowchart dataset described in~\cite{Awal:2011}.
The dataset includes 7 symbol classes (\emph{arrow},
\emph{connection}, \emph{data}, \emph{decision}, \emph{process},
\emph{terminator}, and \emph{text}), and three relation classes
(\emph{Src}, \emph{Targ}, and \emph{AssTxt}). An example was presented
in Section~\ref{sec:related} (Fig.~\ref{fig:fc_input}), with strokes
colored according to the symbol type they belong to. In this dataset,
relations in each flowchart are established between ``adjacent''
symbols. For instance, in the flowchart of Figure~\ref{fig:fc_input},
\emph{Src} and
\emph{Targ} relations are defined between the top \emph{arrow} and a
\emph{terminal} and \emph{data}, respectively. In the same way, an
\emph{AssTxt} relation is defined between the top \emph{terminal} and
the \emph{text} inside it. The flowcharts have been written by $36$
people, and the dataset is divided into a train set with 248 and a
test set with 171 flowcharts. The total number of symbols is about
$9,000$.

As described in Section~\ref{sec:related}, \emph{text} symbols 
have different characteristics than other flowchart symbols, 
and they are usually recognized through specific methods. Since
flowchart recognition is addressed in this work with the aim of
illustrating the application of the proposed framework, we are not
specially concerned with recognition performance. Thus, we have opted
on removing strokes corresponding to text symbols, as well as the
respective relations (\emph{AssTxt}) from the flowcharts. Symbol class
\emph{text} and relation class \emph{AssTxt} were not considered. We
note, however, that it would be equally possible to parse the integral
flowchart without any changes in the parsing and tree extraction steps
once adequate symbol and relation classifiers are developed for texts.

In contrast to the CROHME-2014 dataset, we found no grammar defined
for the flowchart dataset. Thus, we defined a grammar with 16 production
rules, where six of them generate the terminal symbols. 
The embedding is defined in a similar way to the one defined for
mathematical expressions, except for the set of edges to be added. Let
$u$ denote the vertex to be replaced in $G$ and $v_i \in V_{G_r}$ the
vertices in the replacing graph. The edges to be added are defined by:
\begin{align}
 \epsilon = &\{(u',v) \,|\, (u',u) \in E_G \mbox{ and } v = \argmin_{v_i} cost_r (u', v_i) \}\,  \cup \nonumber \\ 
 & \{(v,u') \,|\, (u,u') \in E_{G} \mbox{ and }  v = \argmin_{v_i} cost_r (v_i, u')\} \nonumber 
\label{eq:embedding2}
\end{align} 
where $cost_r(u, v)$ is the minimum relation cost
among relations between a symbol hypothesis under $u$ and a symbol
hypothesis under $v$.


\subsubsection{Parameter adjustment}

For symbol segmentation and classification we used the same method
used for mathematical expressions. Symbol and relation classifier
thresholds, $t_{\mathrm{symb}}$ and $t_{\mathrm{rel}}$, were set both
to $0.95$, following the same scheme as done with mathematical
expressions.

An important performance difference between the two applications is
the relative low accuracy of the flowchart relation classifier
compared to the mathematical expression relation classifier. This
difference is due to the fact that arrows in flowcharts present a high
shape variance and the classifiers we used, which are mainly based on
shape histograms of the symbols~\cite{Frank:2016}, do not generalize
well. We alleviate this deficiency by setting $t_{\mathrm{rel}}=0.95$
(in mathematical expressions, we set $t_{\mathrm{rel}}=0.85$), in
order to keep more labels. We also applied the pruning method based on
the mean junk score of groups with five or more symbols hypotheses,
with $t_{\mathrm{junk}}=0.25$ (see Section~\ref{sec:pruning}) to cope
with the large number valid partitions. For tree
extraction, best validation results were achieved with $\alpha=0.8$,
placing more weight to symbol classifier scores than to relation
classifier scores, and tree pruning threshold $t_{\mathrm{pr}}=0.1$
(same value as in the case of mathematical expressions).

\section{Results and discussions}
\label{sec:experimentation}

Using the datasets, grammars and parameters as described in the
previous section, we applied the recognizers on the test set of the
respective applications. Here we present and discuss the results. 

\subsection{Recognition of mathematical expressions}

Table~\ref{tab:expression_results_test} shows expression level
recognition rates including those reported in the CROHME-2014
competition~\cite{crohme:2014}. The competing systems are identified
as I, $\ldots$, VII, as reported in the competition results.  
The four error columns indicate recognition rates considering
recognition with up to 0, 1, 2 and 3 errors, respectively,

\begin{table}[h]
\caption{Expression level recognition rates on the test set of
  CROHME-2014 competition: competing systems and our method}
\centering
\begin{tabular}{lcccc}
\cline{1-5}
\multirow{2}{*}{\textbf{System}} &  \multicolumn{4}{c}{\textbf{$\#$ errors}} \\ \cline{2-5} 
 & \textbf{0} &  \textbf{$\leq 1$ }&  \textbf{$\leq 2$}&\textbf{$\leq 3$} \\ \hline \hline
I & 37.22 & 44.22 & 47.26 & 50.20 \\
II & 15.01 & 22.31 & 26.57 & 27.69 \\
III & 62.68 & 72.31 & 75.15 & 76.88 \\
IV & 18.97 & 28.19 & 32.35 & 33.37 \\
V & 18.97 & 26.37 & 30.83 & 32.96 \\
VI &25.66 & 33.16 & 35.90 & 37.32 \\
VII & 26.06 & 33.87 & 38.54 & 39.96 \\ \hline
Ours & 33.98 & 43.10 & 47.56 & 49.29\\ \hline
\end{tabular}
\label{tab:expression_results_test}
\end{table}

Our method recognized 33.98\% of the expressions completely. We note,
however, that $78.40\%$ of the generated hypotheses graph include the
complete expressions. Thus, we conclude that the tree extraction
process is failing in retrieving the correct interpretation. This
observation is also consistent with the evaluation performed on the
validation set and described in the previous section.

The two best systems, I and III, include statistical
models~\cite{crohme:2014}. In particular, system III corresponds to
the commercial system
\textit{MyScript}\footnote{http://www.myscript.com/}, which has been
optimized over hundreds of thousands of equations that are not
publicly available. The statistical information used by both systems
could explain, at some extent, their better
performance. Nevertheless, our method is very close in performance to
system I, the best one among those trained exclusively with
CROHME-2014 dataset.

\begin{figure}[h!]
 \centering

 \subfloat[$2^2{b_2} + 2{b_1} + b_0 \rightarrow 2^2{b_2} + 2{b_1} + b0$]{\includegraphics[width=0.6\linewidth]{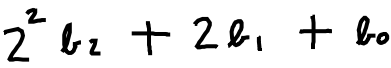}}
   \hspace{20mm}
 \subfloat[$n - n _ {1} - \ldots - n _ {p _ {- 1}} \rightarrow n - n _ {1} - \ldots - n _ {p - 1}$]
 {\includegraphics[width=0.45\linewidth]{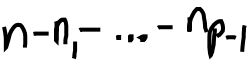} \label{fig:wrong_math_interpretations_c}}
 \hspace{10mm}
  \subfloat[$bag_1 \rightarrow bay1$]{\includegraphics[width=0.24\linewidth]{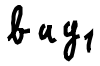}}
  \hspace{20mm}
 \subfloat[$a_0+3a_1+9a_2+27a_3=0 \rightarrow a_0+3a_1+ga_2+27a_3=0$]{ \includegraphics[width=0.8\linewidth]{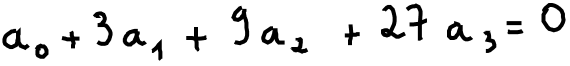}}
 
\caption{Expressions recognized with a few errors. For each expression, its ground 
truth and the system's output is shown as: ground truth $\rightarrow$ system's output.}
\label{fig:wrong_math_interpretations}
\end{figure}

We also note that about $15\%$ of the expressions were not correctly
recognized due to up to 3
errors. Figure~\ref{fig:wrong_math_interpretations} shows some of the
expressions that fall in this case. For instance, in the first
example, the last term $b_0$ was recognized as $b0$. In the last
example, a $9$ is mistaken as $g$. Thus, we hypothesize that several
of the errors could be eliminated by improving the symbol and relation
classifiers. However, some cases are difficult to solve even for
humans. For instance, in the second example, the relation between $p$
and $-1$ is interpreted as \textit{horizontal} and recognized as
$p-1$, when the true relation is \textit{subscript} ($p_{-1}$).

\begin{figure}[h]
 \centering
 \subfloat[]{\includegraphics[width=0.4\linewidth]{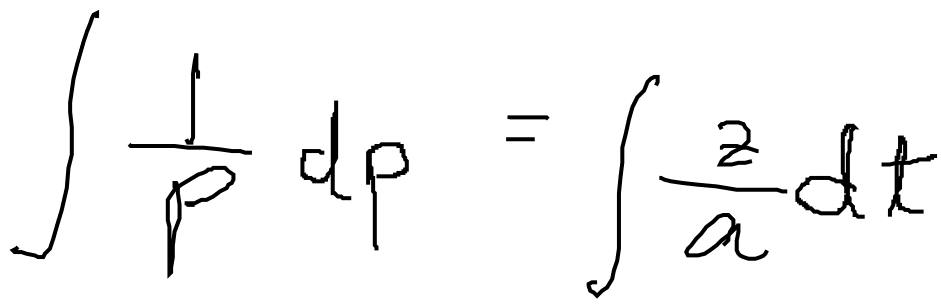}}
 \hspace{10mm}
 \subfloat[]{\includegraphics[width=0.42\linewidth]{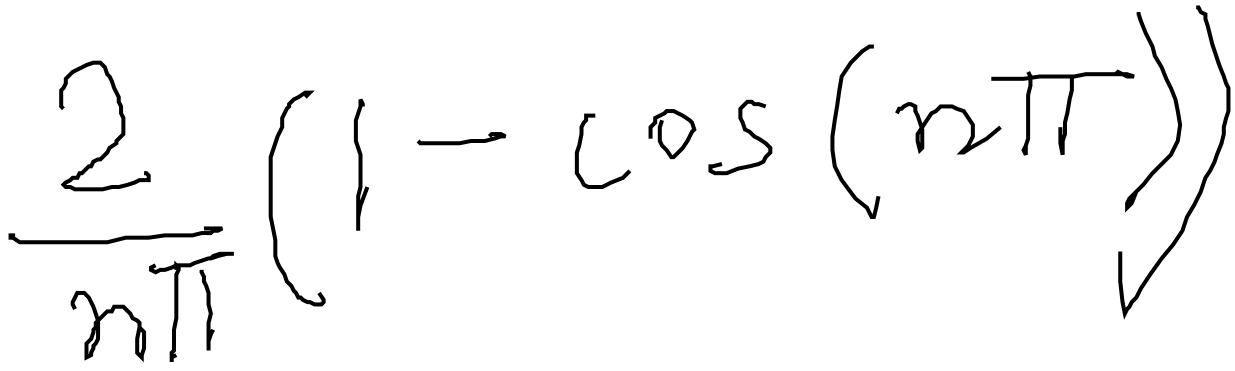}}
   \hspace{10mm}
 \subfloat[]{\label{fig:correct_math_interpretations_c} \includegraphics[width=0.42\linewidth]{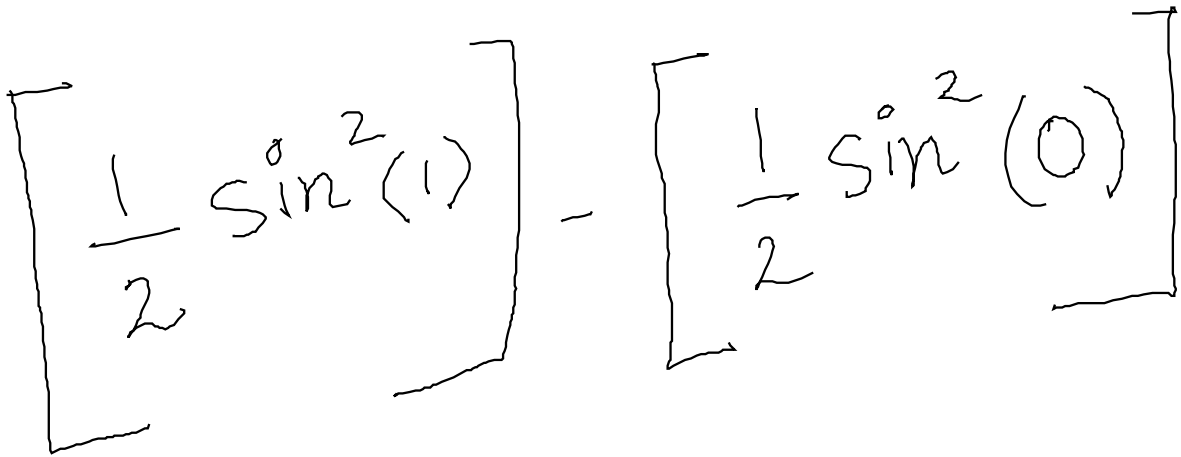}}
 \hspace{10mm}
 \subfloat[]{ \includegraphics[width=0.35\linewidth]{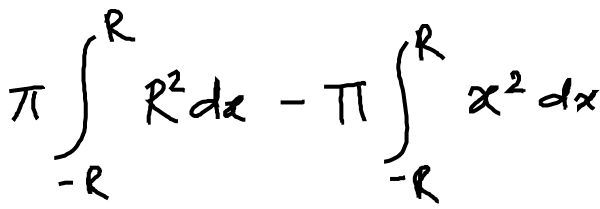}}
 \hspace{10mm}
 \subfloat[]{\includegraphics[width=0.42\linewidth]{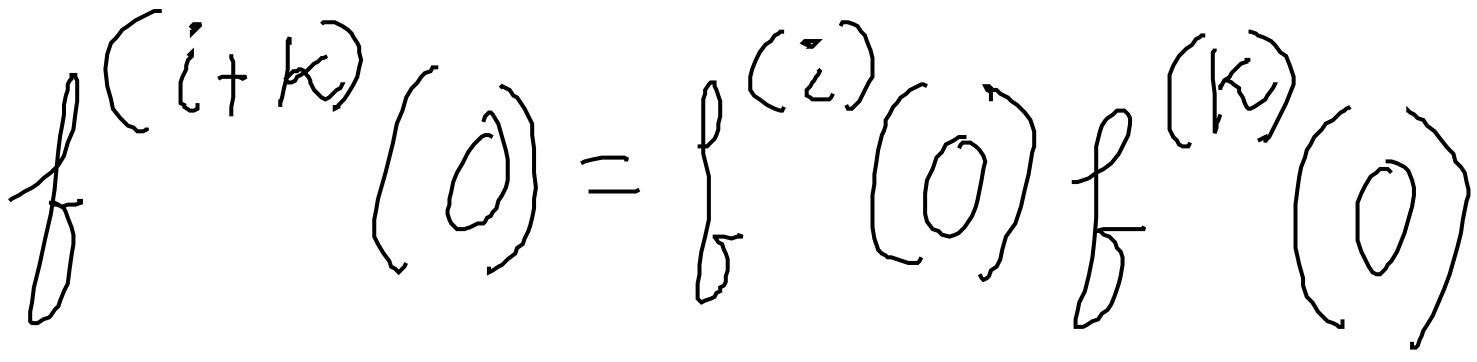}}
  \hspace{10mm}
 \subfloat[]{ \includegraphics[width=0.43\linewidth]{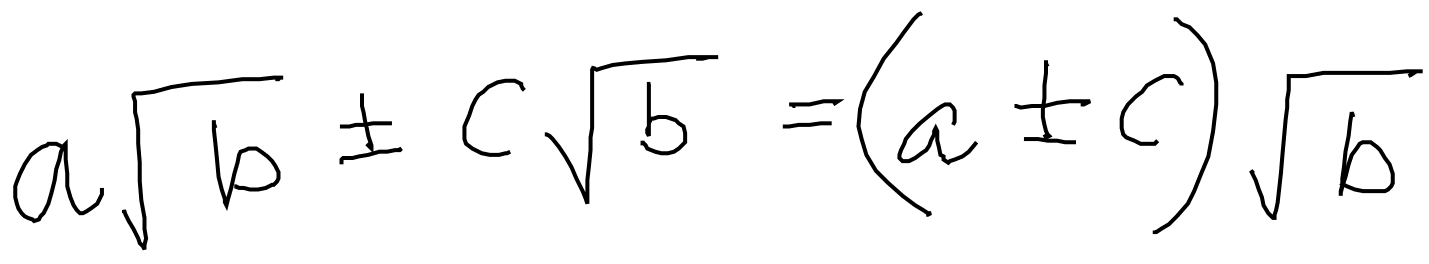}}
 
\caption{Examples of expressions containing potentially ambiguous
  interpretations that have been correctly recognized
  by our system.}
\label{fig:correct_math_interpretations}
\end{figure}

Figure~\ref{fig:correct_math_interpretations} shows examples of
correctly recognized expressions.
Our method is able to correctly recognize some ambiguous symbols as
well as relations.
For instance, in spite of the relation between the subexpressions
``$\frac{1}{2}$'' and ``$\sin^{2}(1)$'' of
Figure~\ref{fig:correct_math_interpretations_c} had received
higher score as \textit{superscript}, the optimal parse tree
interpreted it correctly as a \textit{horizontal} relation.

We also analyzed the most common symbol-to-symbol relation
classification errors on test set.
A classification was considered an error if either the relation or one
of the symbols were wrongly identified. Table~\ref{tab:bigram_errors}
shows the ten most frequent errors. Some of the structures are
particularly difficult due to the ambiguity at symbol level. For
instance, our system often missrecognizes ``$\times$'' as ``$x$'' and
the trigonometric function ``$\sin$'' as tree symbols (like ``$s$'',
``$i$'' and ``$n$'') related by \textit{horizontal} relation.

\begin{table}[h]
\caption{Most frequently misclassified spatial relation between
  symbols on test set. Relation identity is implicitly indicated by
  the relative positions of the symbols.}
\begin{center}
\begin{tabular}{|c|c|c|c|}
\hline
Relation & \# errors & \# samples & $\%$ errors\\ \hline 
$x \times$ &  24 & 24 & 100\\
$\times x$ &  24 & 24 & 100\\
$\sqrt{-}$ &  19 & 29 & 65.52\\
$\displaystyle \frac{}{n}$ &  18 & 37 & 48.65\\ 
$\sin ($ &  20 & 42 & 47.62\\
$=-$ &  26 & 91 & 28.57\\
$\displaystyle \frac{}{2}$ &  19 & 81 & 23.46 \\
$x+$ &  26 & 120 & 21.67 \\
$\frac{1}{}$ &  28 & 133 & 21.05\\
$( x$ &  21 & 108 & 19.44\\\hline
\end{tabular}
\end{center}
\label{tab:bigram_errors}
\end{table}

In mathematical expressions, the probability of certain symbols or
structures be present in particular subexpressions might help solving
ambiguities that can not be solved based only on shapes, relations or
time related information. For instance, the above common errors of
missrecognizing symbol ``$\times$'' as ``$x$'' or the trigonometric
function ``$\sin$'' as tree separated symbols are examples that could
be adequately handled with a statistical model. In the first case,
symbol ``$\times$'' probably appears more frequently between a pair of
numbers (or letters) and probably almost never without two
\textit{arguments} (one at its left side and another at its right
side); in the second case, the three symbols would probably appear
more often as the trigonometric function ``$\sin$'', rather than for
instance, representing the product of three variables $s$, $i$ and $n$.
Hence, statistical information calculated from training data could be
associated to the production rules of the grammar and then rule
probabilities could be considered when ranking the parse trees, by
including a new term in the cost function.


\subsection{Recognition of flowcharts}

Table~\ref{tab:fc_grained_results_test} shows the parsing results
regarding stroke and symbol labeling accuracy w.r.t. the flowchart
test set. It should be noted, however, that we as well as Bresler
\emph{et al.}~\cite{Bresler:2013} did not consider \emph{text}
symbols. 

\begin{table}[h]
\caption{Comparison of our method and four state-of-the-art methods,
  w.r.t. stroke and symbol labeling accuracy ($\%$)}
\centering
\begin{tabular}{lcc}
\cline{1-3}
 \multicolumn{1}{c}{\textbf{System}} & \textbf{Stroke labeling} & \textbf{Symbol labeling} \\ \hline  \hline
Include text recognition: & & \\
Lemaitre \emph{et al.}~\cite{Lemaitre:2013} & 91.1  & 72.4  \\
Carton \emph{et al.}~\cite{Carton:2013}& 92.4  & 75.0  \\ 
Bresler \emph{et al.}~\cite{Bresler:2014} & 95.2  & 82.8  \\ 
Wang \emph{et al.}~\cite{Wang:2016} & 95.8  & 84.3  \\ \hline \hline
Without text recognition: & & \\
Bresler \emph{et al.}~\cite{Bresler:2013} & -  & 74.3  \\ 
Ours & 91.1 & 85.5  \\ \hline
\end{tabular}
\label{tab:fc_grained_results_test}
\end{table}

\begin{figure}[h]
 \centering
 \subfloat[]{\label{fig:fc_26_7b} \includegraphics[trim={0 25cm 17cm 10cm},clip, width=0.5\linewidth]{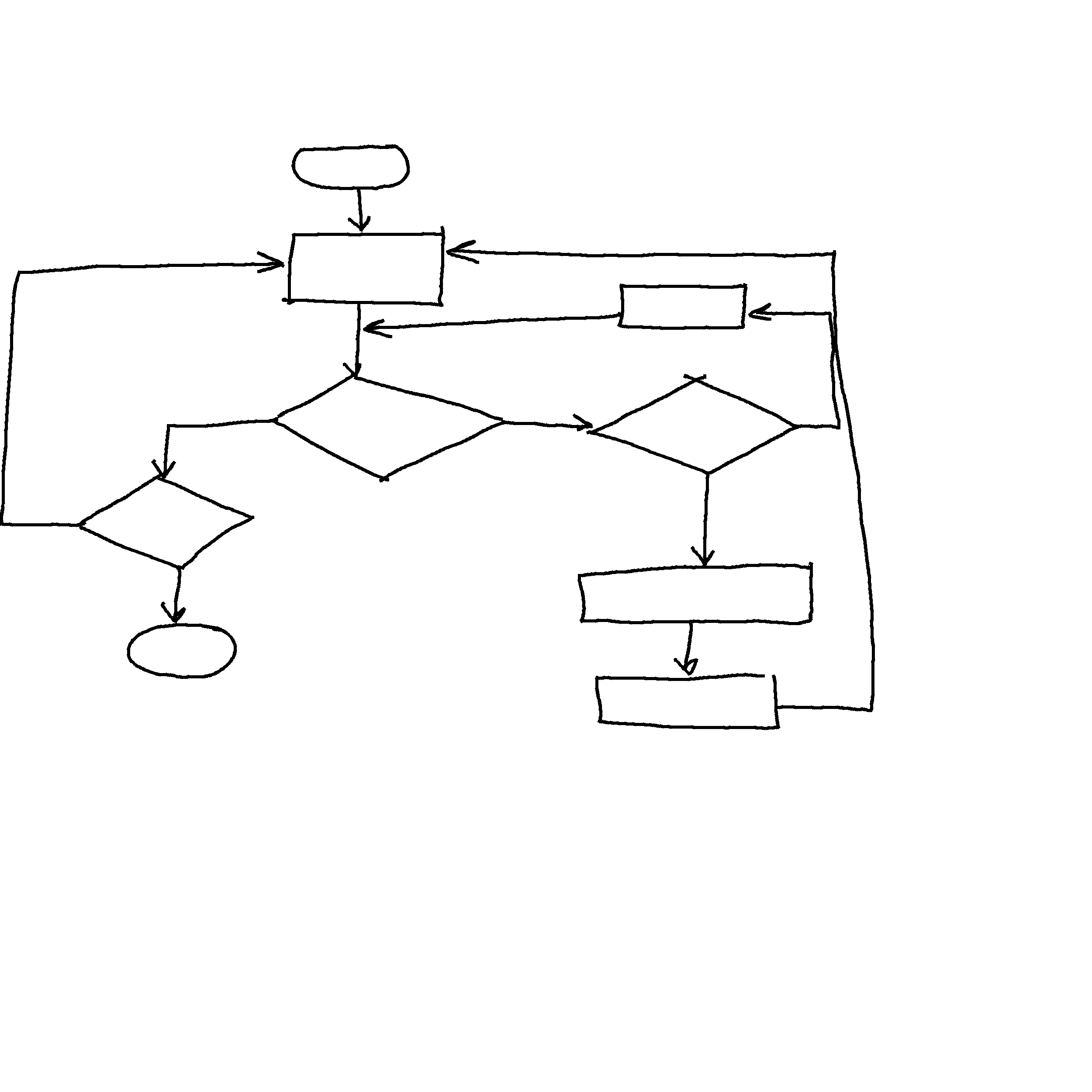}}
\\
 \subfloat[]{\label{fig:fc_12_14b} \includegraphics[trim={0 15cm 24cm 0},clip, width=0.4\linewidth]{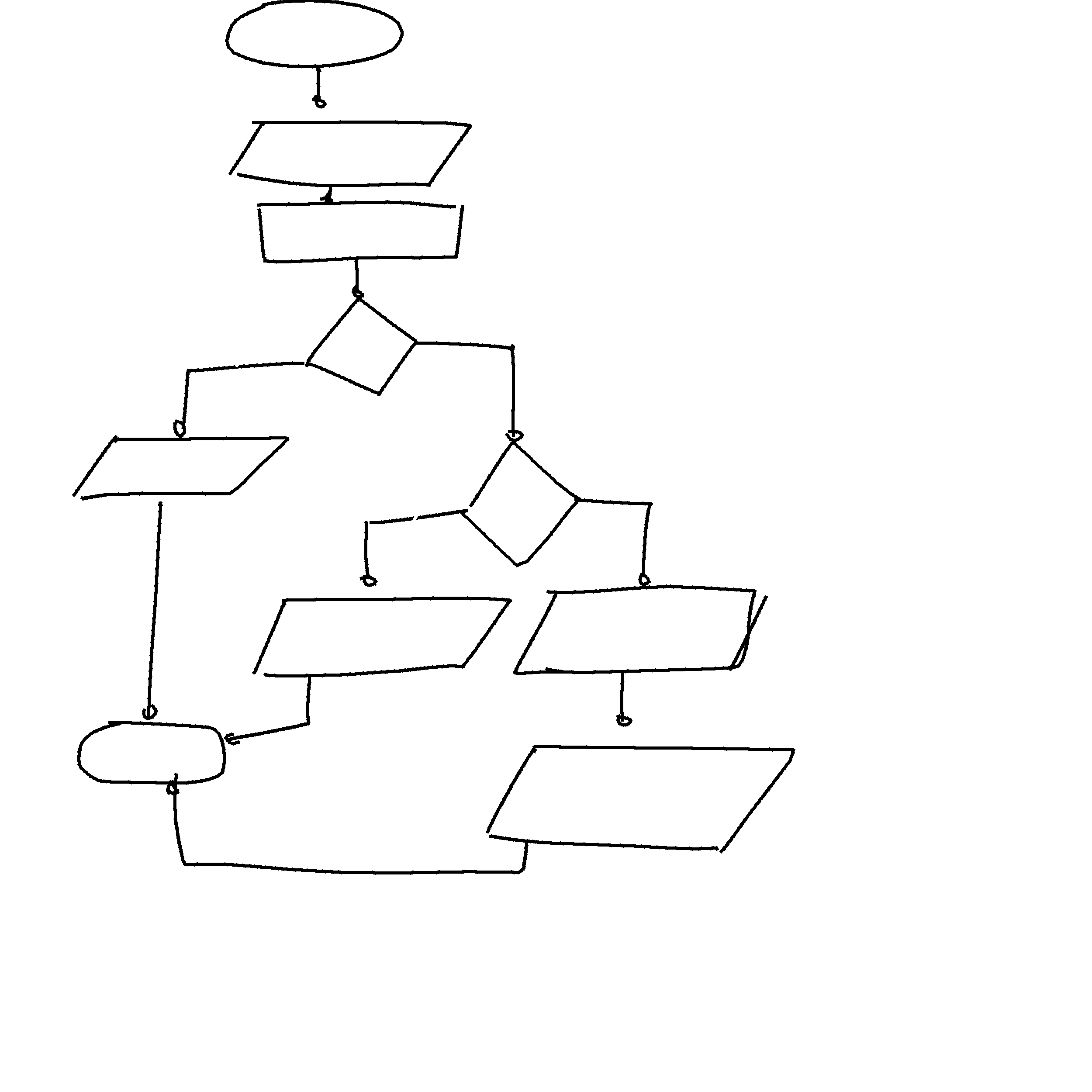}} 
  \hspace{5mm}
  \subfloat[]{\label{fig:fc_16_12b} \includegraphics[trim={0 15cm 22cm 0},clip, width=0.4\linewidth]{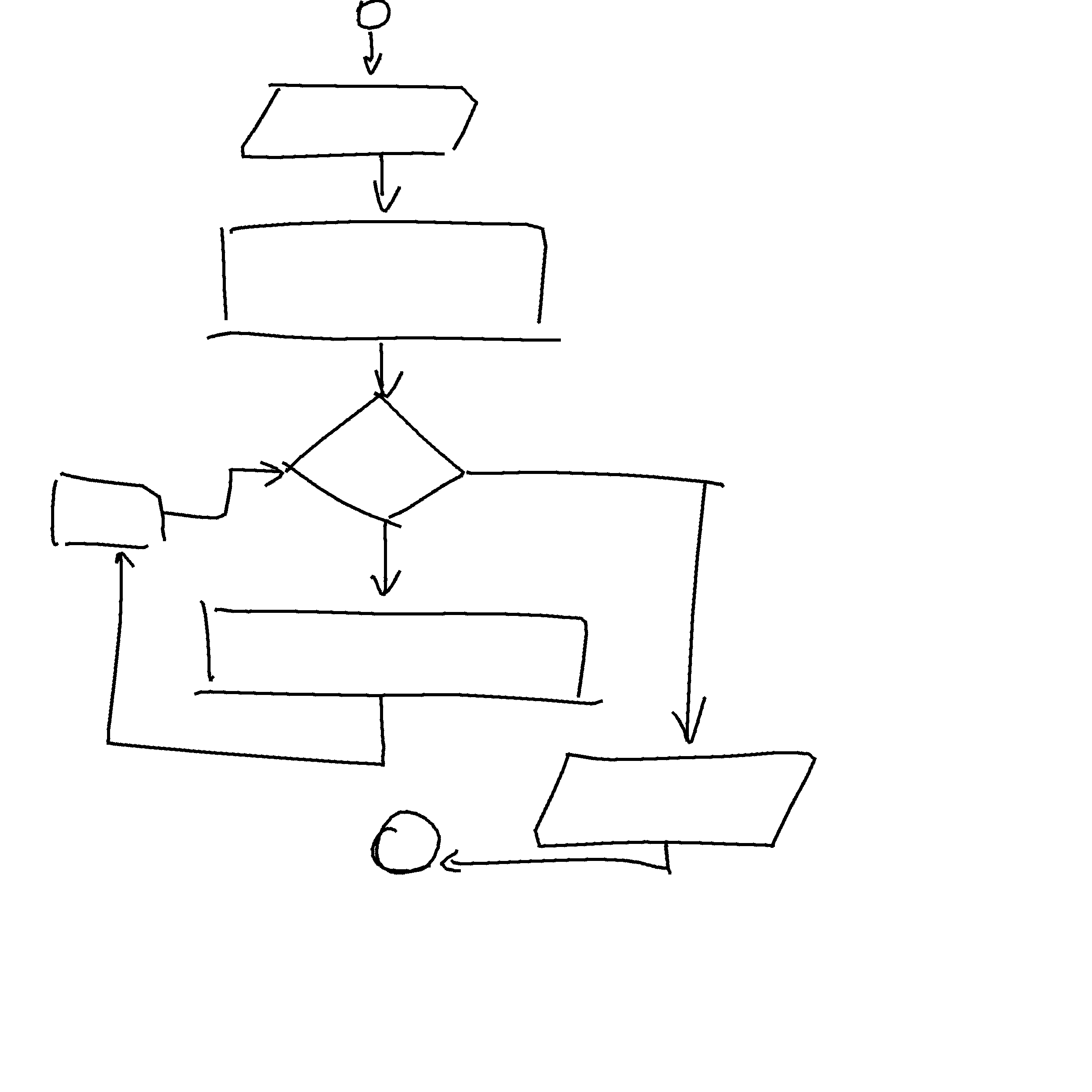}}
\caption{Examples of flowcharts that have been correctly recognized by our method.}
\label{fig:correct_fcs}
\end{figure}

Concerning flowcharts as a whole, our method fully recognized $34\%$
of the flowcharts in the test set. Three examples are shown in
Figure~\ref{fig:correct_fcs}. They include linear as well as (nested)
loop structures. It is interesting to note that varying shapes of
arrows such as the one that extends over a large part of the flowchart
in the right side of Figure~\ref{fig:fc_26_7b}, or very short ones, or
yet curvy ones, are correctly identified.

\begin{figure}[h] 
 \centering 
 \subfloat[]{\label{fig:fc_error_a}
   \includegraphics[width=0.3\linewidth]{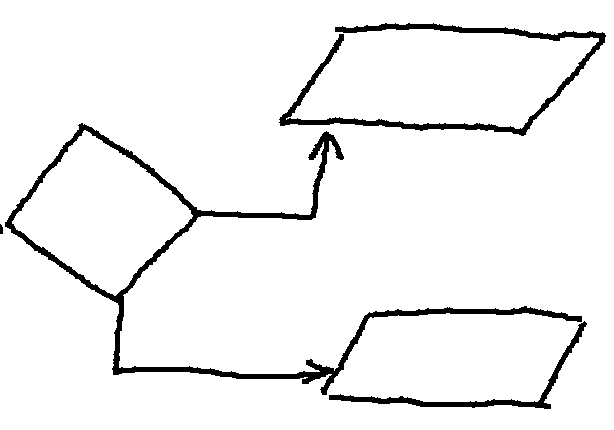}}
\hspace{10mm}
 \subfloat[]{ \includegraphics[width=0.35\linewidth]{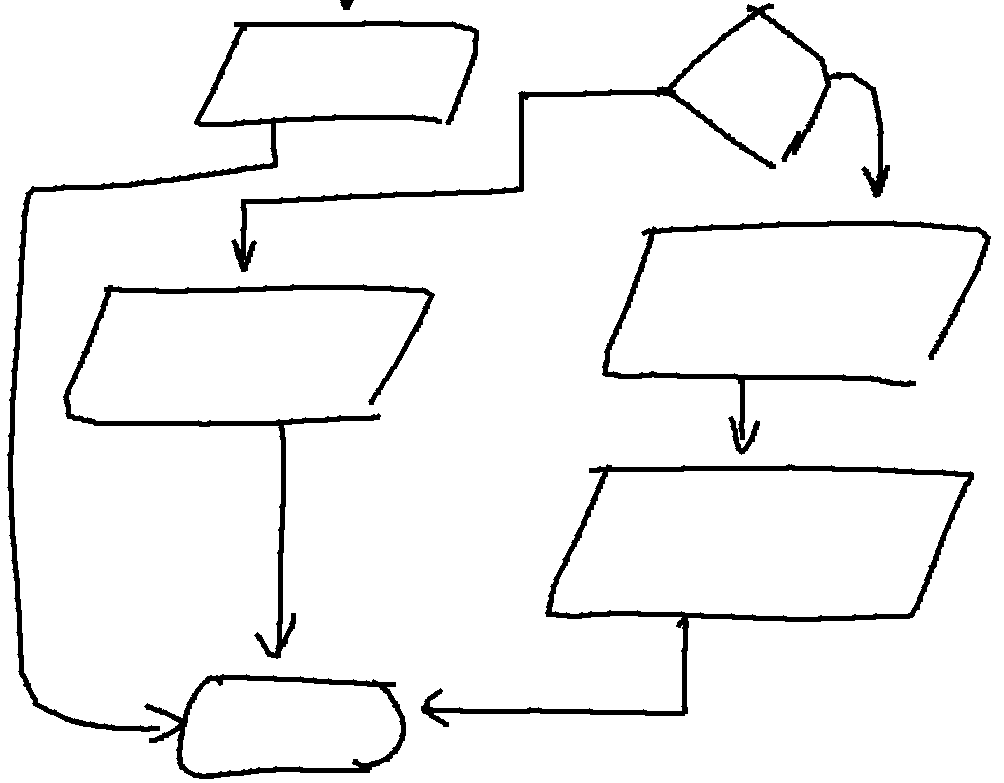}} 
\caption{Parts of flowcharts with missing components in the hypotheses 
graph. (a) Relation between the top arrow and data symbols has not
been identified; (b) the top-center arrow has not been identified.} 
\label{fig:fc_errors}
\end{figure}

When a true symbol or a true relation is not in the hypotheses graph,
the parsing process will fail to recognize the
graphic. Figure~\ref{fig:fc_errors} shows an example of spatial
relation and another of a symbol that were not recognized during the
hypotheses graph generation.

However, since $67\%$ of the flowcharts were fully represented in
the hypotheses graphs, there is a gap of $33\%$ between the achieved
rate and the potentially achievable one. Our explanation for this gap
is the fact that although a relatively large number of symbols are
correctly recognized (Table~\ref{tab:fc_grained_results_test}), many
of the true labels in the hypotheses graph presented lower likelihood
scores than the false ones, leading to a wrong choice of a tree.
Regarding this issue, it is worth to mention that most of the compared
methods used specific techniques to identify flowchart symbols, while
we used a generic method.



The results indicate, nonetheless, that our method can be applied to
flowchart recognition as well. To improve recognition performance, the
current bottleneck seems to be in the hypotheses graph generation
step. By improving symbol and relation classifiers, a considerable
improvement would be possible.

\section{Conclusions}
\label{sec:conclusions}

We have proposed a general framework for the recognition of online
handwritten graphics that is flexible with respect to the family of
graphics, offers possibilities to control processing time, and
integrates symbol and structural level information
in the parsing process. We model graphics as labeled graphs, and the
recognition problem as a graph parsing problem guided by a graph
grammar. The first step of the framework builds a hypotheses
graph that encodes symbol and relation hypotheses computed from the
input strokes. The second step parses the set of strokes
according to a graph grammar. Rule application is modeled as graph
matching between graphs in the rule and graphs induced by partitions
of the stroke set. The parsing step typically generates multiple
interpretations and thus the third step is for selecting an optimal
interpretation. The recognition process is modeled as a
bottom-up/top-down approach, where the hypotheses graph relates to the
bottom-up part that deals with symbol level information and the graph
grammar relates to the top-down part that deals with structural
information.

Flexibility with respect to application domains is achieved by
encoding all domain specific information in the hypotheses
graph and in the grammar, making the parsing method be independent of
a particular application. We presented applications of the framework to
the recognition of mathematical expressions and flowcharts. Recognition
performance are on par with many state-of-the-art methods. Moreover,
our evaluations show that there is room for significative
improvement. Specifically, in mathematical expression recognition we
verified that although $78\%$ of the test expressions were fully
represented in the hypotheses graph, only $33.98\%$ of the expressions
were fully recognized, corresponding to a gap of almost $45\%$. Since
the parsing algorithm generates all interpretations that are
consistent with the grammar, we conclude that the tree extraction step
is failing in choosing the correct interpretation. With respect to
flowcharts, in many cases the  true symbol and relation labels
presented very low likelihood or were not even included in the
hypotheses graph (it should be noted that no specialized symbol or
relation classifier was developed for this application). These
evaluations suggest that an
immediate improvement would be possible by just improving symbol and
relation classifiers. With respect to optimal tree selection,
improvement of symbol and hypotheses likelihood scores will naturally
lead to better cost estimation. However, a second improvement could be
possible by incorporating in the cost computation a term that captures
statistical information with respect to structure occurrence.

Another important feature of our framework is the possibility of
managing computational cost. Hypotheses graph is the main tool to
reduce the space of partitions to be examined when applying a
rule. Only partitions that are present in the hypotheses graph are
considered. In addition, there is a set of parameters to control the
amount of possibilities to be encoded in the hypotheses graph (symbol
and relation label pruning), as well as the number of tree
(interpretation) costs to be evaluated (tree pruning).
These parameters can be adjusted according to each application
particularities.

As future works, we would like to experiment deep neural networks as
tools to improve symbol and relation classification in both
applications and verify how far recognition rate can be pushed.
We would like also to extend the applications to other
families of graphics or 2D structures.

\ifCLASSOPTIONcompsoc
  \section*{Acknowledgments}
\else
  \section*{Acknowledgment}
\fi

This work has received support from CNPq, Brazil (grant 484572/2013-0).
F. Julca-Aguilar thanks FAPESP, Brazil, for the financial support
(2012/08389-1 and 2013/13535-0). N.S.T. Hirata is partially supported
by CNPq (grant 305055/2015-1).

\ifCLASSOPTIONcaptionsoff
  \newpage
\fi

\bibliographystyle{IEEEtran}
\bibliography{bibliography}


\begin{IEEEbiographynophoto}{Frank Julca-Aguilar}
is a postdoctoral researcher at University of S\~{a}o 
Paulo (Brazil). He received his B.Sc degree in Computer Science from 
National University of Trujillo (Peru), and his 
PhD degree in Computer Science from 
University of Nantes (France) and University of S\~{a}o Paulo.
His research interests include machine learning and 
graph-based methods applied to computer vision, 
handwritten recognition, and image processing. 
\end{IEEEbiographynophoto}

\begin{IEEEbiographynophoto}{Harold Mouch\`{e}re}
received his Ph.D. degree in Computer Science from INSA in Rennes, 
France, in 2007 and is now Associate Professor at University of Nantes, France. 
After four years at the IRISA laboratory, he integrated in 2008 the IRCCyN laboratory 
which became in 2017 the LS2N (\textit{Laboratoire des Sciences du Num\'erique de Nantes}). 
His research concerns Pattern Recognition and Machine Learning with application 
to structured document analysis (handwritten mathematical expression, 
on-line flowchart and ancient document analysis). Since 2011, 
he is in the organization committee of CROHME competitions.
\end{IEEEbiographynophoto}

\begin{IEEEbiographynophoto}{Christian Viard-Gaudin}
is a Full Professor at the Electrical and Electronic Engineering 
Department of University of Nantes. He is a leading researcher in the field of
document image processing and handwriting recognition. He has been
involved in many projects concerning automatic mail sorting systems,
offline and online handwriting recognition.
Currently, he is working on mathematical expression recognition,
writer identification and document categorization. 
\end{IEEEbiographynophoto}

\begin{IEEEbiographynophoto}{Nina S. T. Hirata}
holds a PhD degree in Computer Science. She is currently an associate
professor at the Department of Computer Science, Institute of
Mathematics and Statistics of University of S\~ao Paulo. Her research
interests include pattern recognition and
image/signal analysis, using approaches based on machine learning,
graphs, mathematical morphology, and other tools, with applications in
a variety of problems such as document image analysis, graphics
recognition, astronomical and plankton image classification, image
segmentation, object detection in images, among others.
\end{IEEEbiographynophoto}

\end{document}